\documentclass[runningheads]{llncs}

\usepackage[mobile]{eccv}

\usepackage{eccvabbrv}

\usepackage{graphicx}
\usepackage{amsmath}
\usepackage{amssymb}
\usepackage{booktabs}
\usepackage{xcolor}
\usepackage{wrapfig,lipsum,booktabs}

\usepackage[accsupp]{axessibility}  

\usepackage[pagebackref,breaklinks,colorlinks]{hyperref}

\usepackage[capitalize]{cleveref}
\crefname{section}{Sec.}{Secs.}
\Crefname{section}{Section}{Sections}
\Crefname{table}{Table}{Tables}
\crefname{table}{Tab.}{Tabs.}

\definecolor{red}{rgb}{1.0, 0.0, 0.0}

\definecolor{green}{rgb}{0.0, 0.5, 0.0}

\usepackage{orcidlink}

\begin{document}

\title{VF-NeRF: Viewshed Fields for Rigid NeRF Registration}

\author{Leo Segre\inst{1} \and
Shai Avidan\inst{1}}

\authorrunning{L. Segre and S. Avidan}

\institute{
Tel Aviv University\\
\url{https://leosegre.github.io/VF_NeRF/}
}

\maketitle

\begin{abstract}
3D scene registration is a fundamental problem in computer vision that seeks the best 6-DoF alignment between two scenes. This problem was extensively investigated in the case of point clouds and meshes, but there has been relatively limited work regarding Neural Radiance Fields (NeRF). 
In this paper, we consider the problem of rigid registration between two NeRFs when the position of the original cameras is not given. Our key novelty is the introduction of Viewshed Fields (VF), an implicit function that determines, for each 3D point, how likely it is to be viewed by the original cameras. We demonstrate how VF can help in the various stages of NeRF registration, with an extensive evaluation showing that VF-NeRF achieves SOTA results on various datasets with different capturing approaches such as LLFF and Objaverese. Our code will be made publicly available.

\begin{figure}
\centering
\begin{subfigure}[b]{0.24\linewidth}
\frame{\includegraphics[width=\textwidth]{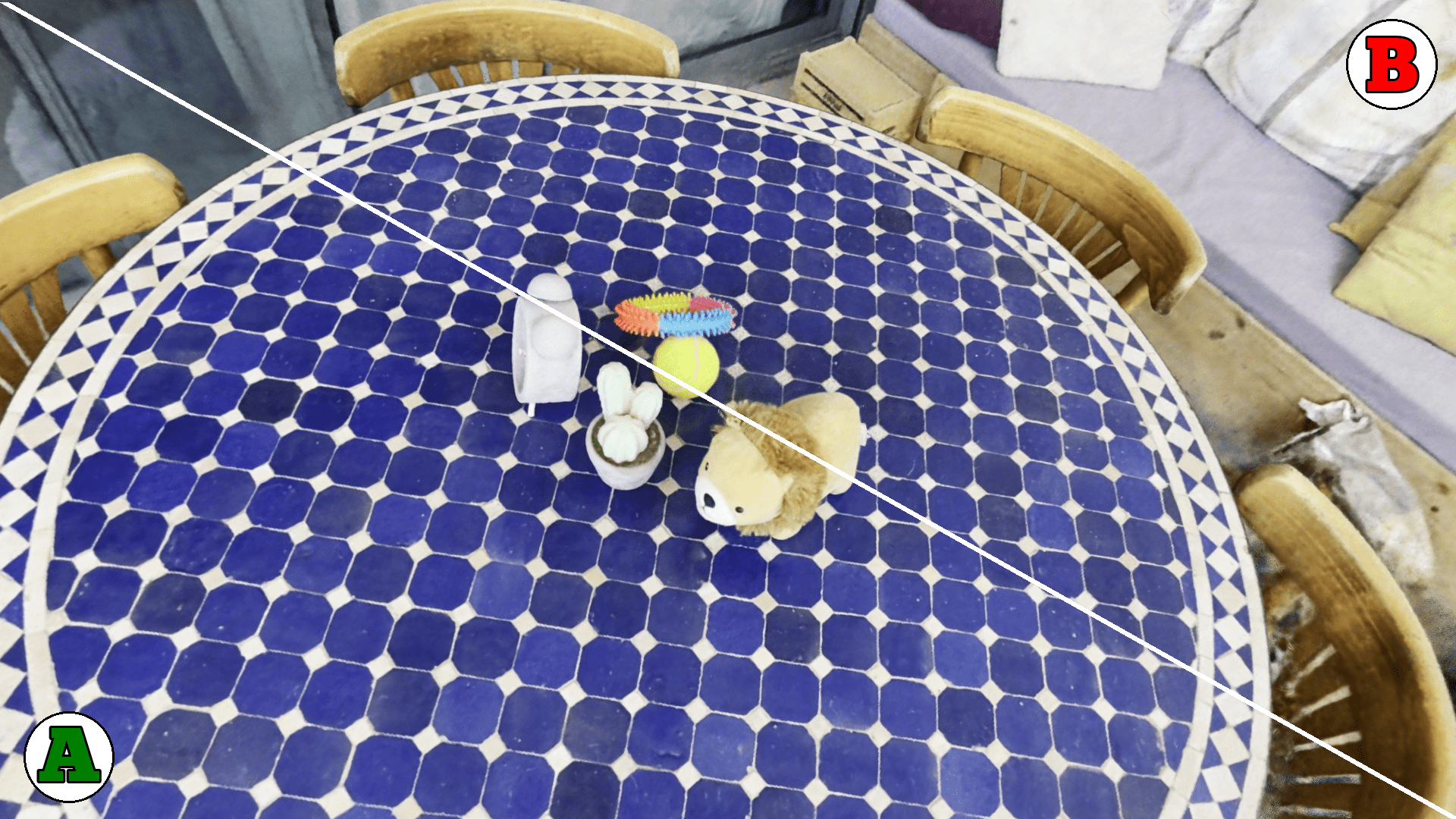}}
\caption{}
\label{fig:merged_table}
\end{subfigure}
\hfill
\begin{subfigure}[b]{0.24\linewidth}
    \frame{\includegraphics[width=\textwidth]{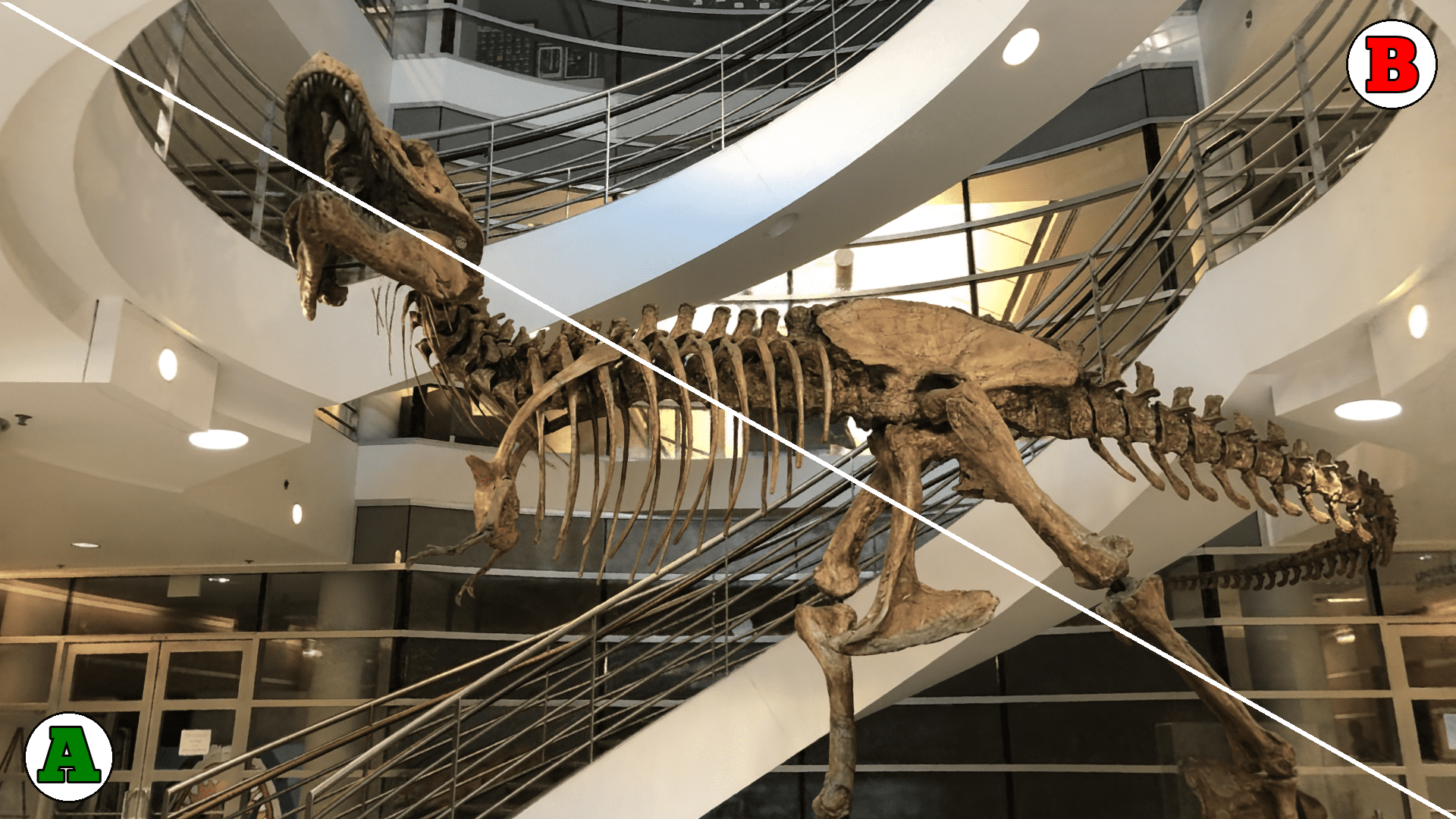}}
    \caption{} 
    \label{fig:merged_trex}
\end{subfigure}
\hfill
    \begin{subfigure}[b]{0.24\linewidth}
    \frame{\includegraphics[width=\textwidth]{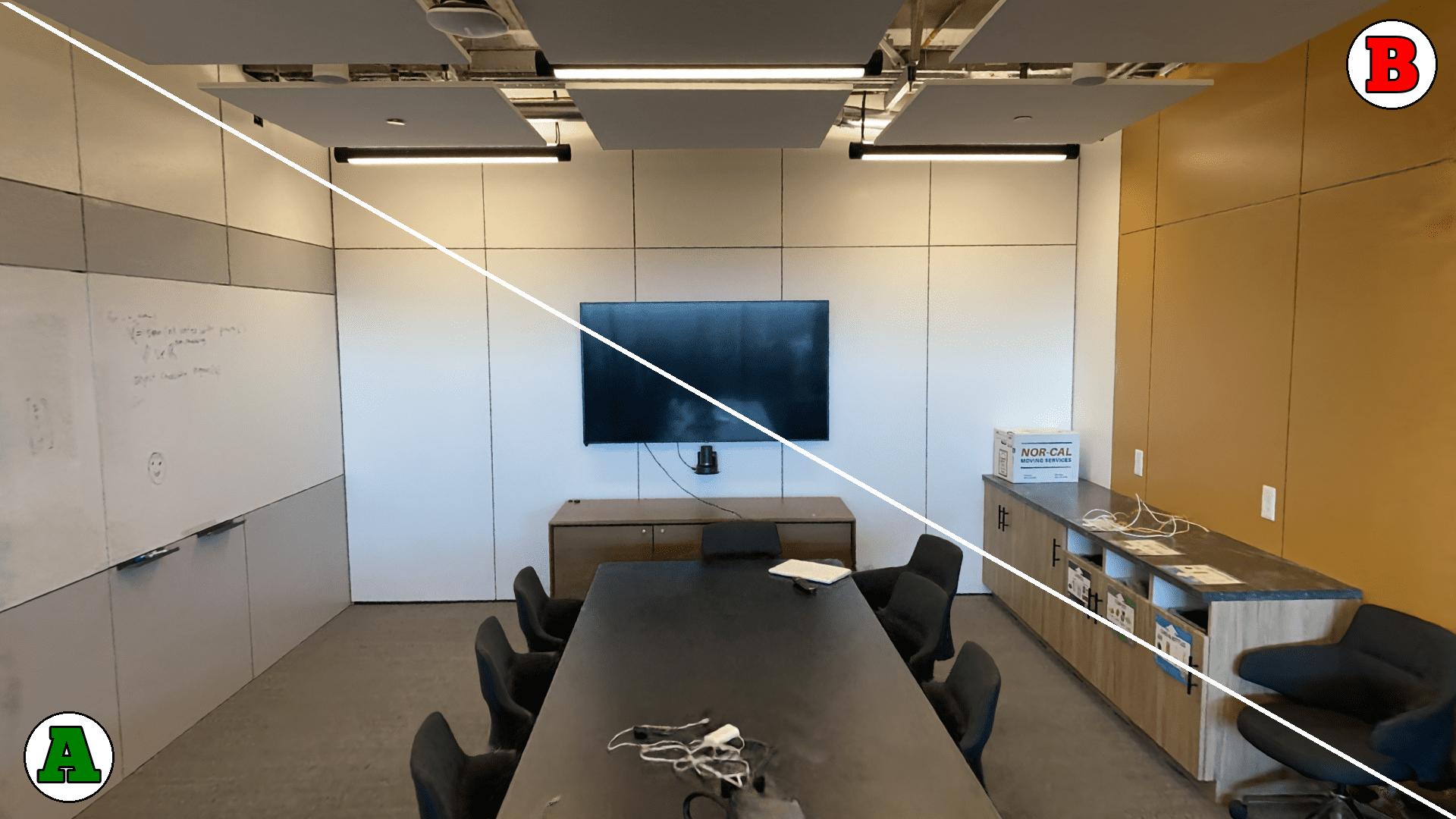}}
    \caption{}
    \label{fig:merged_room}
\end{subfigure}
\hfill
    \begin{subfigure}[b]{0.24\linewidth}
    \frame{\includegraphics[width=\textwidth]{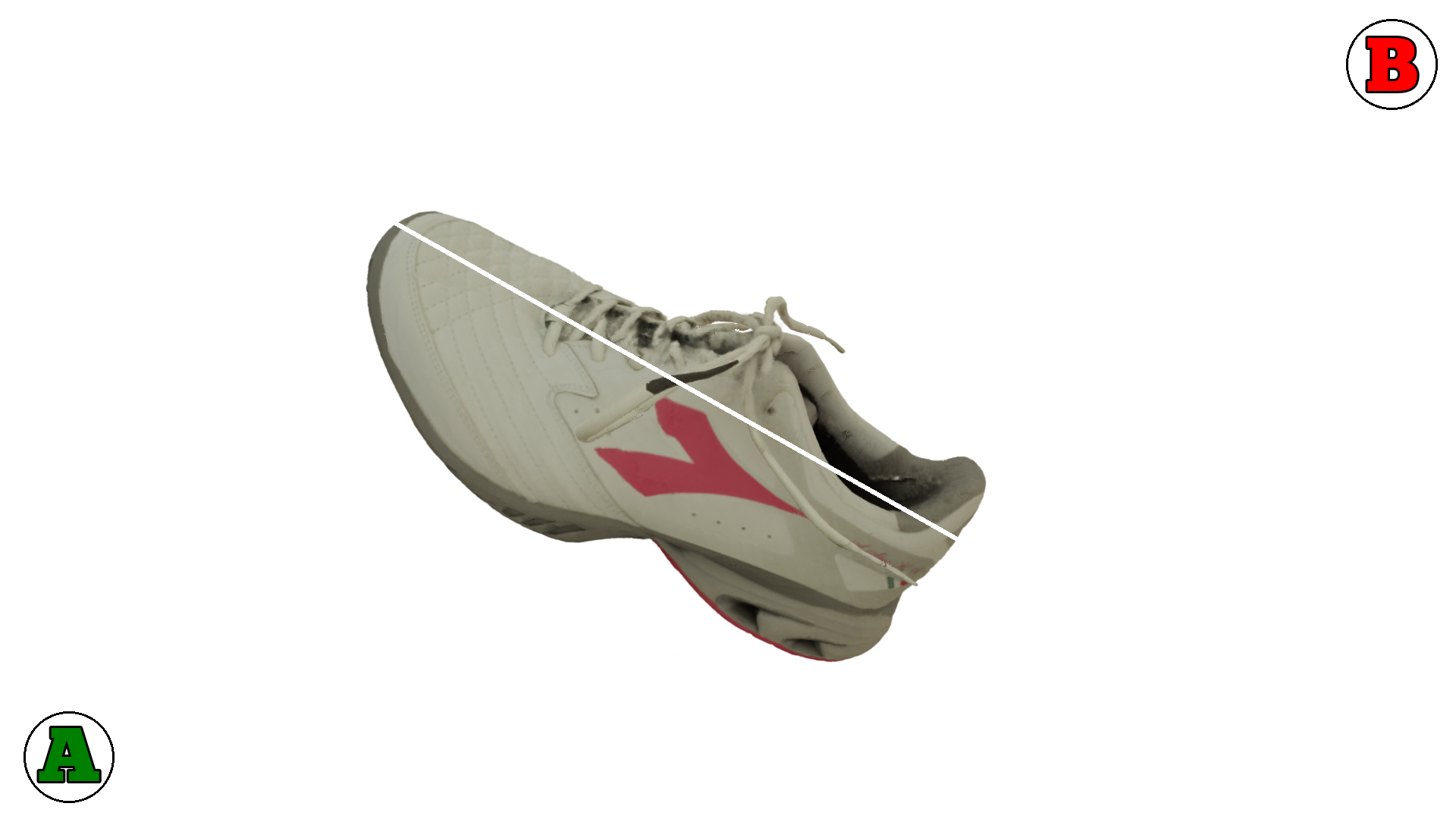}}
    \caption{}
    \label{fig:merged_sneaker}
\end{subfigure}
\caption{\textbf{NeRF Registration:} Registration results of VF-NeRF for four different scenes. All images are from novel view points where the bottom left part is taken from one NeRF and the top right part is taken from the registered second NeRF. \ref{fig:merged_table} is of a casually captured scene, \ref{fig:merged_trex} and \ref{fig:merged_room} are taken from LLFF dataset and \ref{fig:merged_sneaker} is from the Objaverse dataset.}
\label{fig:teaser}
\end{figure}

\end{abstract}

\vspace{-1cm}
\section{Introduction}
Registering two 3D scenes is a fundamental problem that has been studied for years in the field of computer vision. Solving it has a large number of applications in many fields ranging from pure image processing such as medical imaging \cite{8398245} and object detection \cite{Choy_2015_CVPR} to world-scale tasks such as mobile robotics \cite{ROB-035} and autonomous driving \cite{9733274}. Until recently, common ways to represent, and register, 3D scenes were based on point clouds or meshes. Recently, Neural Radiance Fields (NeRF~\cite{10.1145/3503250}) emerged as a viable alternative, and we propose a registration algorithm that operates directly on them.

Our approach is simple and straightforward. Generate a set of images from the source NeRF and seek a rigid transformation that minimizes a photometric loss of these images with respect to the target NeRF. The transformation that minimizes this loss is the one that registers the two NeRFs. 
Figure~\ref{fig:teaser} shows registration results from several different scenes. 

\begin{figure}[t]
\centering
\includegraphics[width=0.9\textwidth]{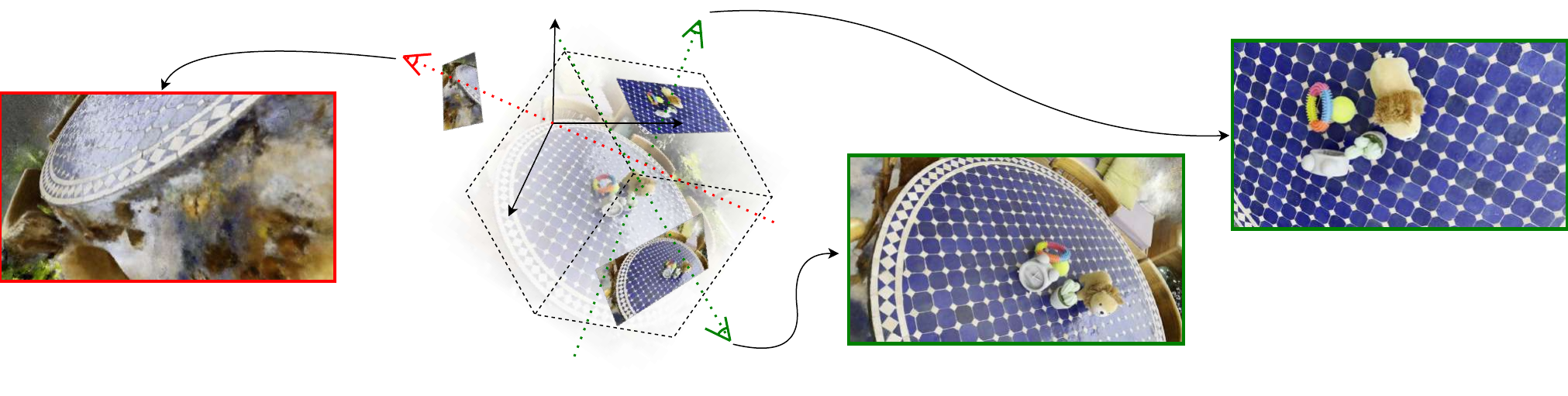}
\hfill
\caption{\textbf{Novel View Generation:} Randomly sampling novel camera parameters often lead to non-informative images. For example, the red-marked image on the left was generated using a camera that lies on the unit sphere looking towards the origin. In contrast, using our novel Viewshed Fields (VF) representation we are able to generate informative camera positions (green marked images on the right) that can then be used to register two NeRFs.}
\label{fig:bad_camera}
\end{figure}

Given only two NeRFs, without the position of the original cameras in either one of them, we are faced with the following question: how to sample "good" virtual camera viewpoints? A possible solution is to sample the position of the virtual camera to be on the unit sphere and point the camera at the origin of the scene. However, as shown in Figure~\ref{fig:bad_camera}, this often leads to poor results, where almost half of the image is essentially noise (red-marked image on the left). Instead of working at the camera level, we aggregate the information of all the original cameras into a novel representation, termed Viewshed Fields (VF), that allows us to generate images like the two green-marked images shown on the right. 

\begin{figure*}[t]
\centering
\includegraphics[width=\textwidth]{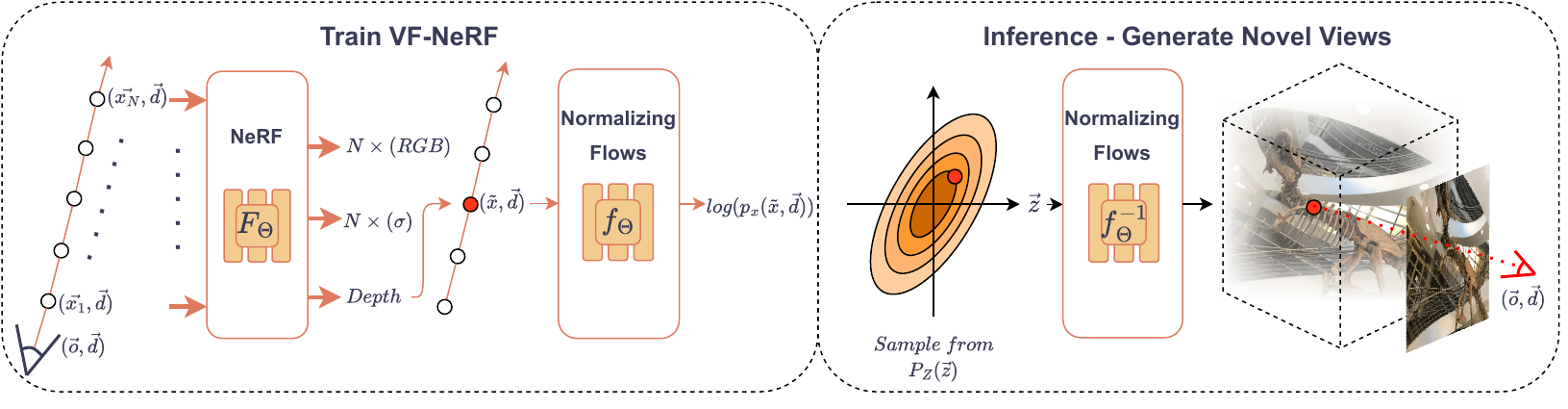}
\caption{\textbf{VF-NeRF:} (Left) Our VF-NeRF consists of two parts. The first is a NeRF network with the standard RGB and $\sigma$ outputs with depth estimation. The second part is a simple normalizing-flows network, where its input is a point on the surface and the camera direction (i.e., an {\em oriented point}) and its output is the log-likelihood estimation that is maximized during the training phase. (Right) To generate novel views we sample from the 6-dimensional Gaussian in the Normalizing-Flows latent space. Then we recover the oriented point $(\Tilde{x}, \Vec{d})$ and use equation \ref{eq:origin_reconsruction} to reconstruct the camera origin. Finally, we render the view of the camera in position $\vec{o}$ and direction $\vec{d}$.}
\label{fig:VF_NeRF}
\end{figure*}

VF is an implicit function, similar to NeRF, that, given an oriented point (\ie, a 3D point and a viewing direction), outputs a scalar that represents how well was the 3D point covered, from a specific direction, by the original set of images that was used to create the NeRF. If we had access to such an oriented point, we could have placed a virtual camera that is looking at it. Unfortunately, VF is an implicit function, and we do not have such access. To overcome this challenge, we treat the problem as a generative process, where the goal is to generate high VF score points. Specifically, we use Normalizing Flows (NF)~\cite{dinh2014nice} to map high value VF points to a Gaussian distribution in latent space during the original NeRF training. Then, we use the generative process to generate (\ie, sample) high value VF points and direct the virtual camera at them. An overview of our method is shown in Figure~\ref{fig:VF_NeRF}.

We also use VF to help initialize the registration process and to help optimization. Specifically, we can use VF to generate a 3D point cloud and rely on the vast literature of point registration to obtain a good initial alignment. Then, during photometric optimization we rely on VF to select high quality rays that lead to better optization results. To summarize, we make the following contributions:
\begin{itemize}

\item We introduce a novel representation, termed Viewshed Fields (VF), that helps register two NeRFs. VF represents 3D points that were well covered by the initial set of cameras that captured the scene.

\item We generate meaningful novel views to support NeRF registration task. We use a generative method, based on Normalizing Flows, to generate high score VF points. These points are then used to set the parameters of virtual cameras which, in turn, produce images of the scene that are used to solve the NeRF registration problem.

\item We show how to use VF to construct point clouds of a scene.
    
\end{itemize}

\section{Related Work}

\subsection{Neural Radiance Fields}
Common ways to represent 3D scenes include  point clouds, meshes, and voxel grids. Recently, NeRF \cite{10.1145/3503250} became a popular choice for representing 3D scenes. It introduced a differentiable renderer to optimize the 3D scene based on 2D RGB images. Utilizing the differentiable renderer, the 3D scene can be learned through back-propagation, hence the scene can be represented by the weights of a neural network, mainly MLP-based. Numerous works leveraged this approach to improve the scene quality \cite{Barron_2021_ICCV, barron2022mip}, accelerate the learning and rendering time \cite{garbin2021fastnerf, mueller2022instant}, utilize depth supervision \cite{deng2022depth} and render more complex fields such as features and semantics \cite{zhi2021place, vora2021nesf, fu2022panoptic, tschernezki2022neural}.

A large body of work considered the use of NeRF in dynamic setting~\cite{Li_2023_CVPR}. This typically requires the alignment of multiple frames to a canonical coordinate system. This alignment is usually achieved in the form of flow estimation and {\em not} through the estimation of a rigid global transformation, as is done here. Recently, Gaussian Splatting~\cite{kerbl3Dgaussians} was introduced as a powerful alternative to NeRF. However, this differs from our work since Gaussians are explicit models as opposed to the implicit nature of NeRFs.

\subsection{3D Scene Registration}
Registration is a well-studied task in the field of computer vision, with a wide range of works related to explicit representations such as point clouds or meshes. Due to the explicit nature of those representations, registration algorithms can work on a finite set of points or vertices and fit the two sets using classic algorithms \cite{4767965, 121791}. Some iterative algorithms use local refinements to tackle the registration task, such as ICP variants \cite{rusinkiewicz2001efficient, li2020evaluation, zinsser2003refined, chetverikov2002trimmed}, but those algorithms are prone to fail given a bad initialization or partial overlap between the sets. In addition to those, there are classic global methods that first match pairs between the two sets \cite{guo2016comprehensive, tombari2013performance, drost2010model, 5152473} and then use a sparse subset of them for global alignment \cite{zhou2016fast}. Recent works learn the alignment features utilizing deep neural networks \cite{choy2020deep, wang2019deep, hezroni2021deepbbs, zhang20233d}. One method that utilize it for NeRF registration is DReg-NeRF \cite{chen2023dregnerf} that converts the NeRF to voxel grid and train a deep neural network for registration task. It achieves improved results over point cloud registration methods but requires a large training set.

NeRF assumes that camera pose is recovered through Bundle Adjustment, such as COLMAP~\cite{schoenberger2016sfm}, in a pre-processing step. Recent works such as BARF \cite{Lin_2021_ICCV} or L2G-NeRF~\cite{chen2023local} demonstrate that bundle adjustment can be done during NeRF training over photometric loss. This line of work takes images as input and outputs a single consistent NeRF. It does not register two NeRFs, as we do. A work that is closer to us is that of iNeRF \cite{yen2020inerf}.  They do not perform NeRF registration, but they do perform image to NeRF registration which can be used for NeRF registration. Their method is based on back-propagating the photometric loss through the NeRF weights to optimize the camera pose.
NeRF2NeRF~\cite{10160794} is another method that works directly with NeRF representation.
They show an improvement over point cloud registration methods, but require user input at the initialization, which is not needed in our approach.

Our work is not to be confused with implicit Signed Distance Function works~\cite{Park_2019_CVPR, chetverikov2002trimmed} that use a latent code-conditioned feed-forward decoder network, or directly fit a Neural Network to a point cloud to generate the surface of a shape. Our goal is not to reconstruct the shape of surfaces in the scene, but rather generate "good" view points.

\subsection{Normalizing Flows}
Normalizing flows (NF) is a family of invertible models that convert data from a "real-world" distribution to a standard distribution latent space and vice versa. That attribute makes NF a generative model that makes it possible to reconstruct data from the original distribution by sampling the known distribution in latent space.

Real-NVP \cite{dinh2017density} proposed an invertible network that is based on MLP and affine coupling layer. The training process maximizes the log-likelihood using a change of variables. This method performs well on low dimensional data, but struggles in the case of high dimensional data such as images and videos. Moreover, the computational effort is proportional to the dimension of the data.  Glow \cite{kingma2018glow} proposed using invertible $1 \times 1$ convolution and a multi-scale architecture to deal with the input dimensionality. Note that in our case, we adopt Real-NVP due to the low dimensionality of our input ($6D$ oriented points).

There has been work on trying to measure uncertainty in NeRF representations. For example, NeRF-W \cite{Martin-Brualla_2021_CVPR} decompose the scene into static and transient objects (i.e., walking people in static scenes). Other works presented uncertainty as an evaluation metric for novel views \cite{sunderhauf2023density, shen2021stochastic}. These metrics can assist in understanding whether a novel view quality is good or not, but it can not assist in locating the scene and generate high-quality novel views. Recently, Conditional-Flow NeRF \cite{shen2022conditional} introduced an NF-based method to measure uncertainty in NeRF. Although this method uses NF, it maximizes the pixel-color likelihood as a function of the NeRF outputs. Hence, the invertible nature of NF is ineffective when trying to locate the scene and generate a forward-facing novel view.

\section{Method}

\begin{figure}[t]
\centering
    \begin{subfigure}[b]{0.49\linewidth}
        \includegraphics[width=\textwidth]{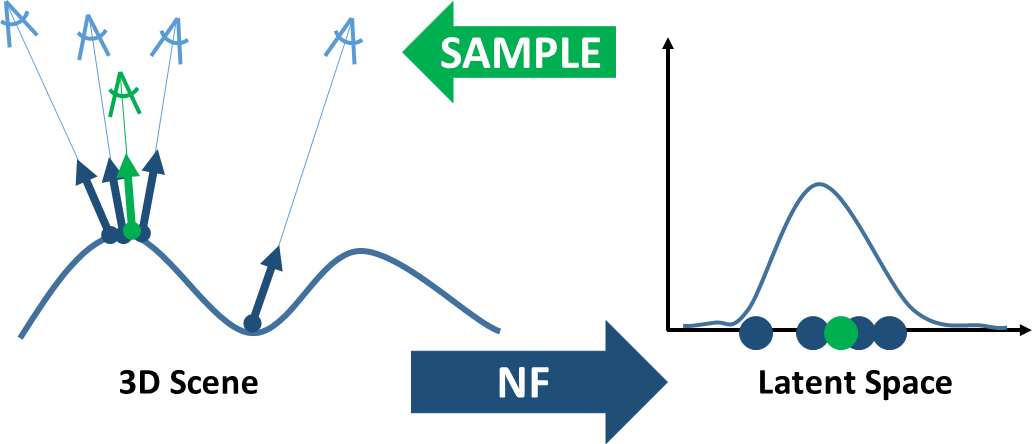}
        \caption{}
        \label{fig:VF}
    \end{subfigure}
    \hfill
    \begin{subfigure}[b]{0.24\linewidth}
        \includegraphics[width=\textwidth]{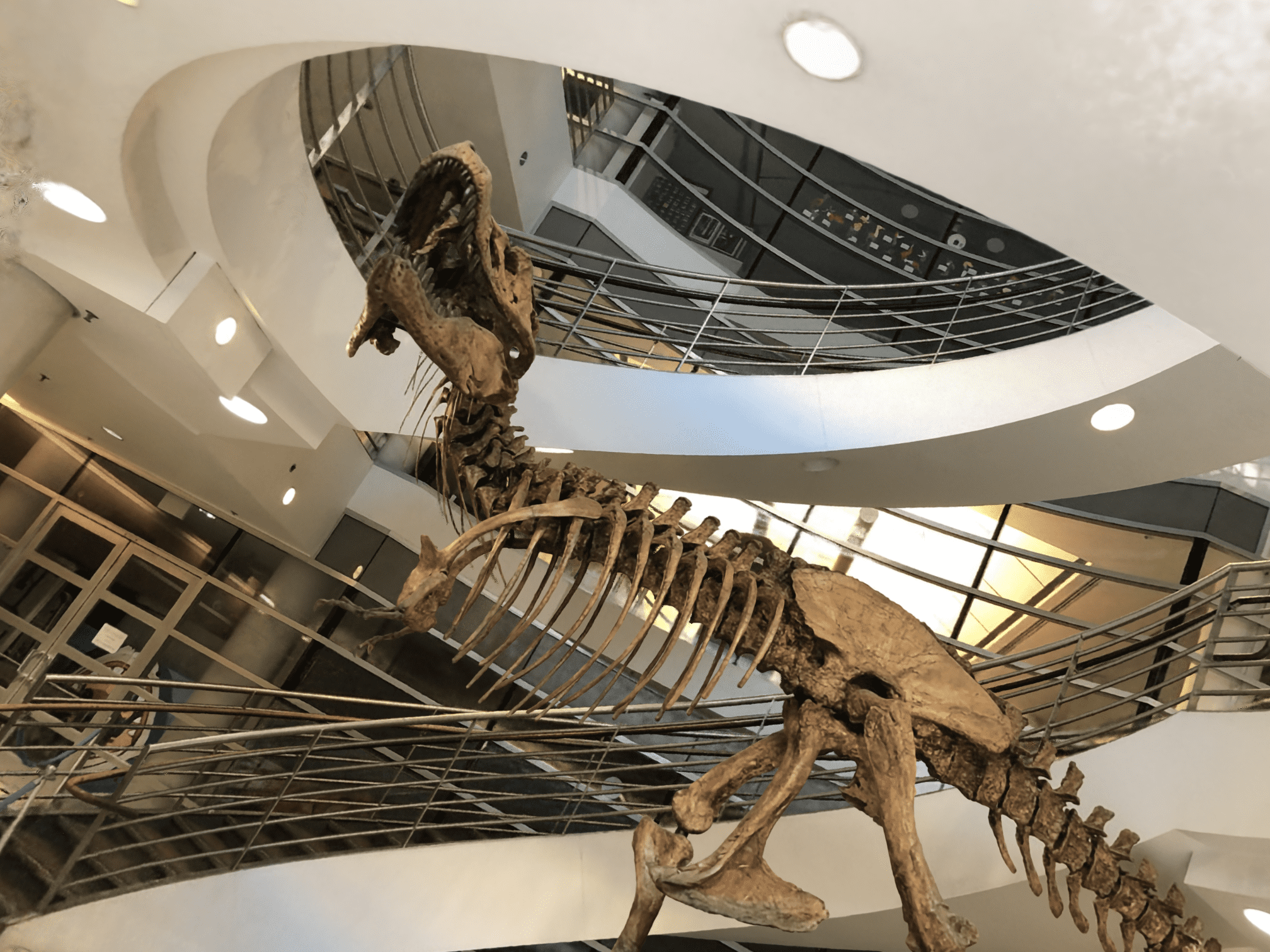}
        \caption{}
        \label{fig:trex_vf_example_rgb}
    \end{subfigure}
    \hfill
    \begin{subfigure}[b]{0.24\linewidth}
        \includegraphics[width=\textwidth]{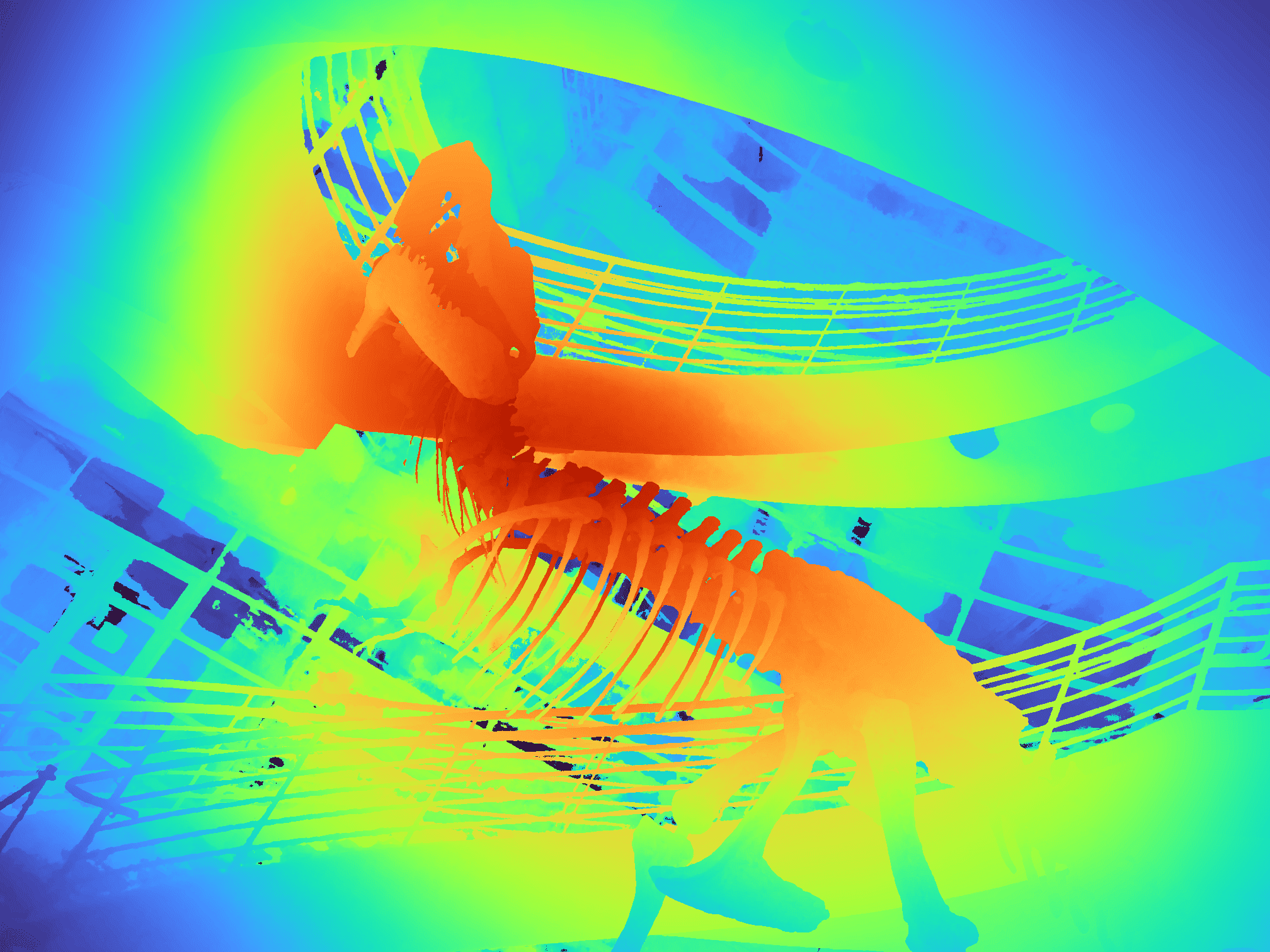}
        \caption{}
        \label{fig:trex_vf_example}
    \end{subfigure}

\caption{\textbf{Viewshed Fields:} (\ref{fig:VF} Left) During NeRF training, we sample {\em oriented points} (blue) around surfaces in the scene and use Normalizing Flows (NF) to map them to a Gaussian in latent space. (\ref{fig:VF} Right) During NeRF registration, we sample a high visibility oriented point (green) from the Gaussian and map it to the input space where it is used to determine the position of the novel camera. \ref{fig:trex_vf_example_rgb} demonstrates a novel view synthesis generated using our method and \ref{fig:trex_vf_example} is the viewshed map generated respectively.}
\label{fig:viewshed}
\end{figure}

We wish to find a 6-DoF transformation between two NeRFs. We do that by generating images from one NeRF and seeking the 6-DoF transformation that will minimize a loss between the images from one NeRF with respect to the other NeRF. Our solution is based on a novel representation, termed Viewshed Fields (VF). See Figure~\ref{fig:viewshed}.

VF works as follows. During NeRF training we obtain {\em oriented points}, which are 3D surface points accompanied by a viewing direction (denoted by blue circles with pointing arrow on the left side of Figure~\ref{fig:VF}). The oriented points are mapped to a Gaussian in latent space using Normalizing Flows (the right part of Figure~\ref{fig:VF}). Once training is done, we can sample the Gaussian in latent space to generate high visibility points (marked in green). These points are points on the surface of objects in the scene that are, with high probability, observed by many cameras. Given the sampled oriented point, we can generate the position and orientation of the novel camera (marked with green in the left part of Figure~\ref{fig:VF}). Figures~\ref{fig:trex_vf_example_rgb} and~\ref{fig:trex_vf_example} show an example novel view, and its corresponding viewshed map, that were generated by sampling VF.

\subsection{Viewshed Fields}

The Viewshed Field is learned during the training phase of VF-NeRF, either together with the density and RGB values, or after the NeRF parameters are fixed (\ie working on a pre-trained NeRF). We use Normalizing-Flows \cite{dinh2014nice} to learn it, where the normalizing-flows model $f: \mathcal{X} \rightarrow \mathcal{Z}$, learns a mapping between the data distribution $\mathcal{X}$ of oriented points $(\Vec{x}, \Vec{d})$ with location $\Vec{x}$ and direction $\Vec{d}$, to a 6 dimensional diagonal Gaussian in latent space $\mathcal{Z}$. We learn the mapping $f$ using Real-NVP~\cite{dinh2017density} architecture with 4 layers. The optimization is done by minimizing the unsupervised negative log-likelihood of $p_X(x)$ through the typical change of variables formula of normalizing flows:
\begin{equation}\label{p_x}
p_x(x) = p_z(f(x)) \cdot \left| \det \left( \frac{df}{dx} \right) \right|
\end{equation}
\begin{equation}\label{log_p}
log(p_x(x)) = log(p_z(f(x))) + \sum_{i=1}^{K} log \left| \det \left( \frac{df_i}{df_{i-1}} \right) \right|
\end{equation}
The data $(\Vec{x},\Vec{d})$ is sampled during  NeRF training, where $\Vec{d}$ is constant along a ray and $\Vec{x}$ varies along it. We sample only the surface of the object along the ray into the Normalizing-flows model. That is, let $\Tilde{x}$ be the single $\Vec{x}$ that lies on the surface of the object and let $\Vec{o}$ be the origin of the ray. The depth is simply defined by the median of the weights (of the densities learned by NeRF) accumulated along the ray.
\begin{equation}\label{x_on_surface}
\Tilde{x} = \Vec{o} + depth \cdot  \Vec{d}
\end{equation}
Now we can sample the pair $(\Tilde{x}, \Vec{d})$ into the normalizing-flows model so the viewshed score (i.e. likelihood of the VF according to the Gaussian) is high on the surface of an object but low everywhere else, as can be seen in figure~\ref{fig:trex_vf_example}.
\paragraph{Novel Views}

Since we have used Normalizing Flows to learn the Viewshed Field, we can sample points from the 6-dimensional Gaussian and map them to oriented points with high viewshed value.
\begin{equation}
\label{eq:x_from_z}
(\vec{x}, \vec{d}) = f^{-1}(\vec{z})
\end{equation}
These oriented points' log probability is formally given by Equation \ref{log_p}, where $f(x) = z$. In practice, we generate $K$ samples from $\mathcal{Z}$, invert them to $K$ oriented points $(\Vec{x}, \Vec{d})$, and finally select the top-$N$ in terms of $log(p_x(x))$. From each oriented point $(\Vec{x}, \Vec{d})$ we can generate a novel view since we know, with high probability, that it is valid to view the point $\Vec{x}$ from direction $\Vec{d}$. Setting the camera origin is done by inverting equation \ref{x_on_surface}.
\begin{equation}
\label{eq:origin_reconsruction}
\Vec{o} = \Vec{x} - depth \cdot  \Vec{d}
\end{equation}
Finally, we render a novel view from the camera with origin $\Vec{o}$ that is facing toward the direction $\Vec{d}$. High viewshed score is not guaranteed for all the pixels in this novel view, so we render the viewshed field and use a threshold to generate a 2D viewshed mask and sample accordingly, as can be seen in Figure \ref{fig:pixel_sampling}. 

\subsection{$\hat{T}_0$ Initialization}\label{initialization}

Let $T \in SE(3)$ be the transformation from scene $A$ to scene $B$. Our approach is based on finding $T$ that minimizes a photometric loss, hence the initialization of $T$ is important. We introduce two different approaches for initialization. Both are based on VF.

\paragraph{Photometric based initialization:}
We use $VF_A$ of scene $A$ to generate a set $C_A = \{C_i | i = 1, 2, 3 ... N\}$ of {\em good} camera view points of scene $A$. Given a possible transformation $T \in SE(3)$, we use $VF_B$ of scene $B$ to determine how well the cameras in $C_A$, that are transformed by $T$, observe {\em good} points in scene $B$. This intuition is captured in the following:
\begin{equation}\label{init_score}
    Score_{init}(C_A; T) = Median_{C_A}(\sum_{p\in T(C_i)} VF_B(p))
\end{equation}
Where $p \in T(C_i)$ refers to all oriented points $p$ that come from camera $C_i$ that underwent transformation $T$. $VF_B(p)$ is the likelihood of the oriented point $p$, according to Normalizing Flows, in scene $B$. For robustness, we take the score of $T$ to be the median over all transformed cameras in the set $C_A$. 
For photometric initialization, we sample, at random, multiple transformations $T$, and pick the one with the highest score.

\paragraph{VF-based point cloud:} The second approach we use is based on converting NeRF to point cloud and relying on the vast literature for point cloud registration. We use VF to sample the point clouds. Specifically, we sample $M$ points from the 6-dimensional Gaussian which is the NF latent distribution and use equation \ref{eq:x_from_z} to reconstruct $M$ oriented points. Then, we sample NeRF with the oriented points as its input to get the corresponding density and RGB. Finally, we utilize density values along with a specified threshold to filter out uncertain points. We currently do not integrate the colors for registration purposes. Once the point clouds of both scenes are generated, we can use any known global registration for point clouds, to serve as our initial guess.

\subsection{Gradient Based Optimization}
\begin{figure}[t]
\centering
\vspace{0pt}
\includegraphics[width=0.32\textwidth]{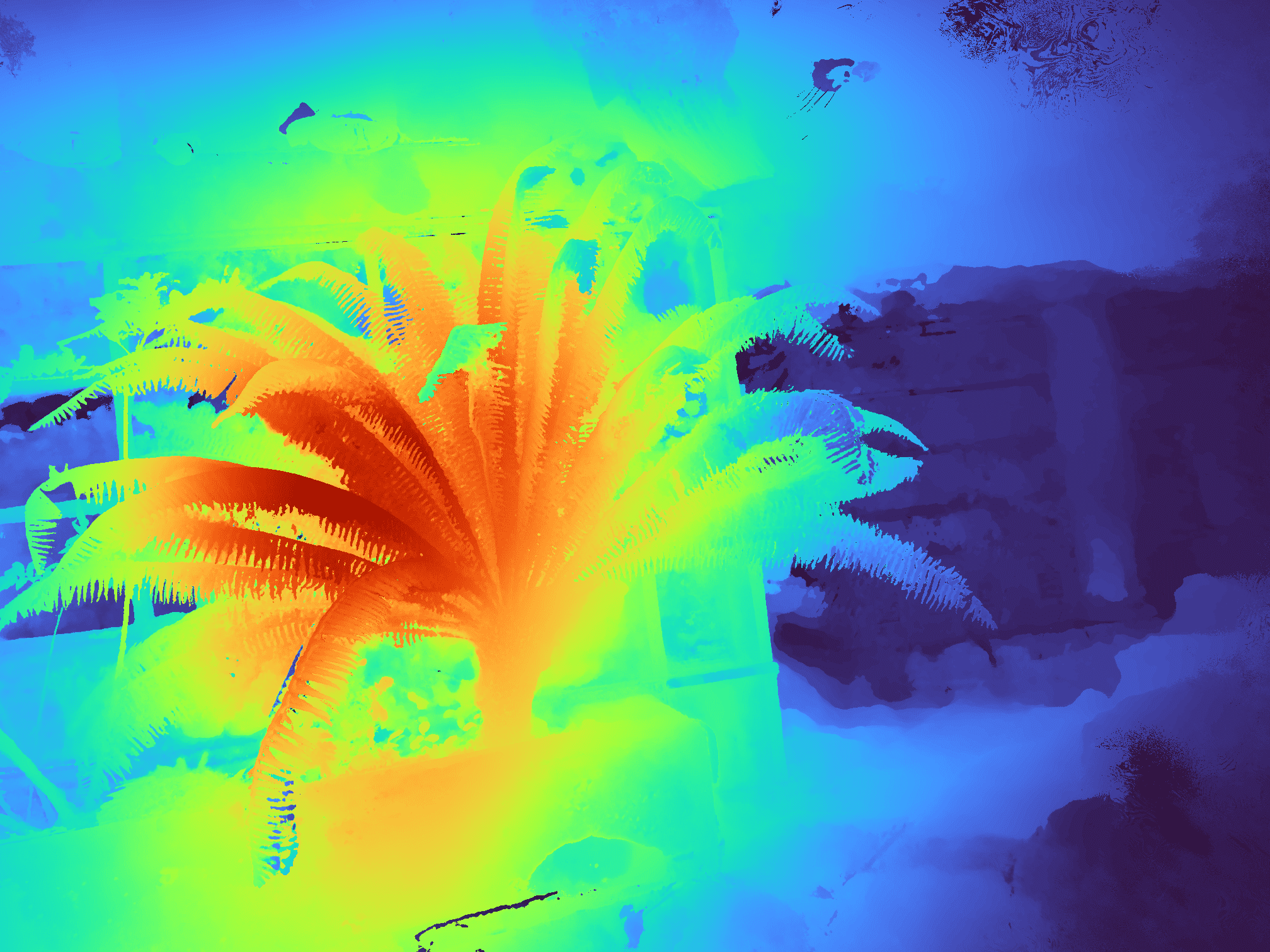}
\includegraphics[width=0.32\textwidth]{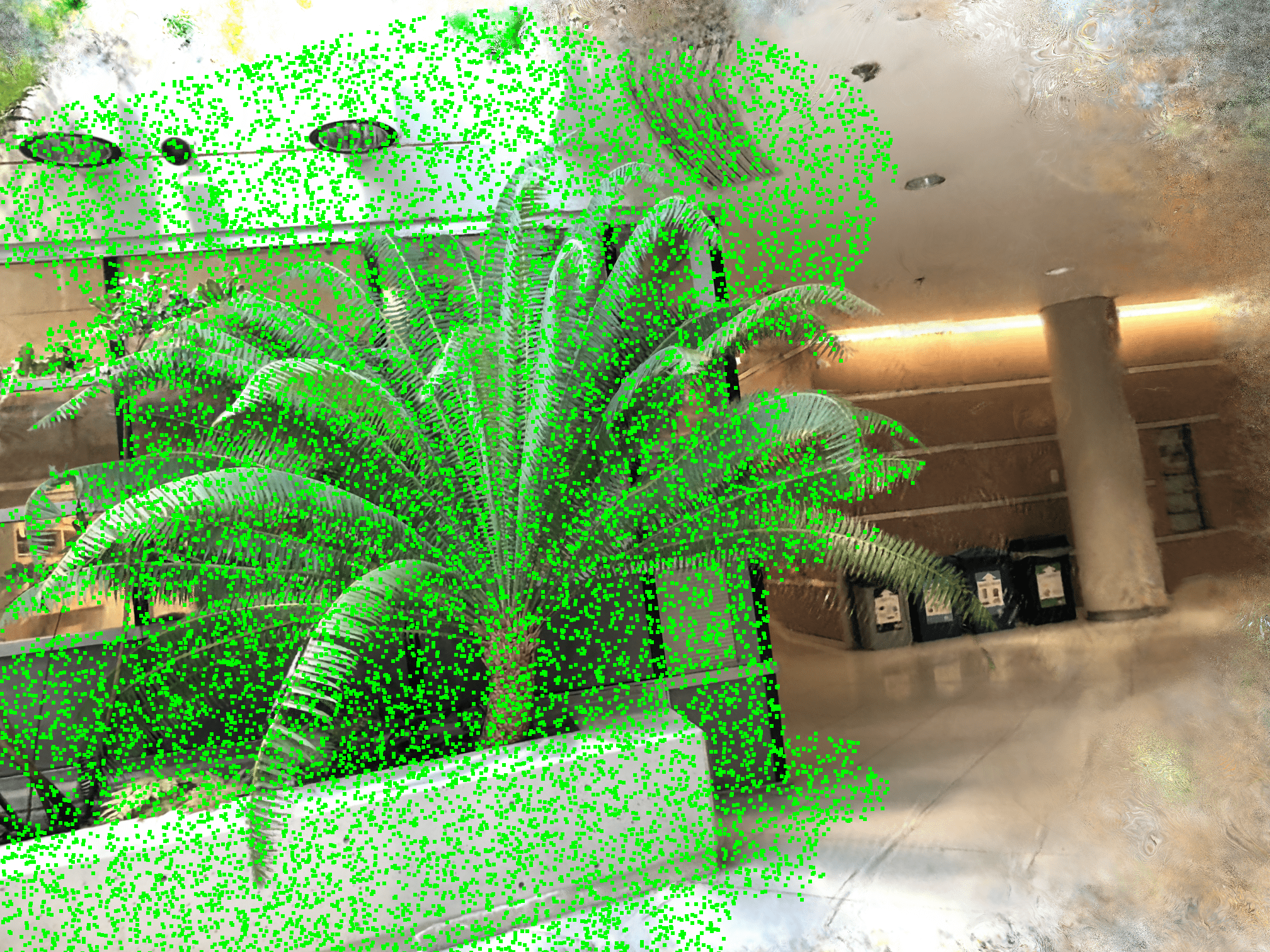}
\includegraphics[width=0.32\textwidth]{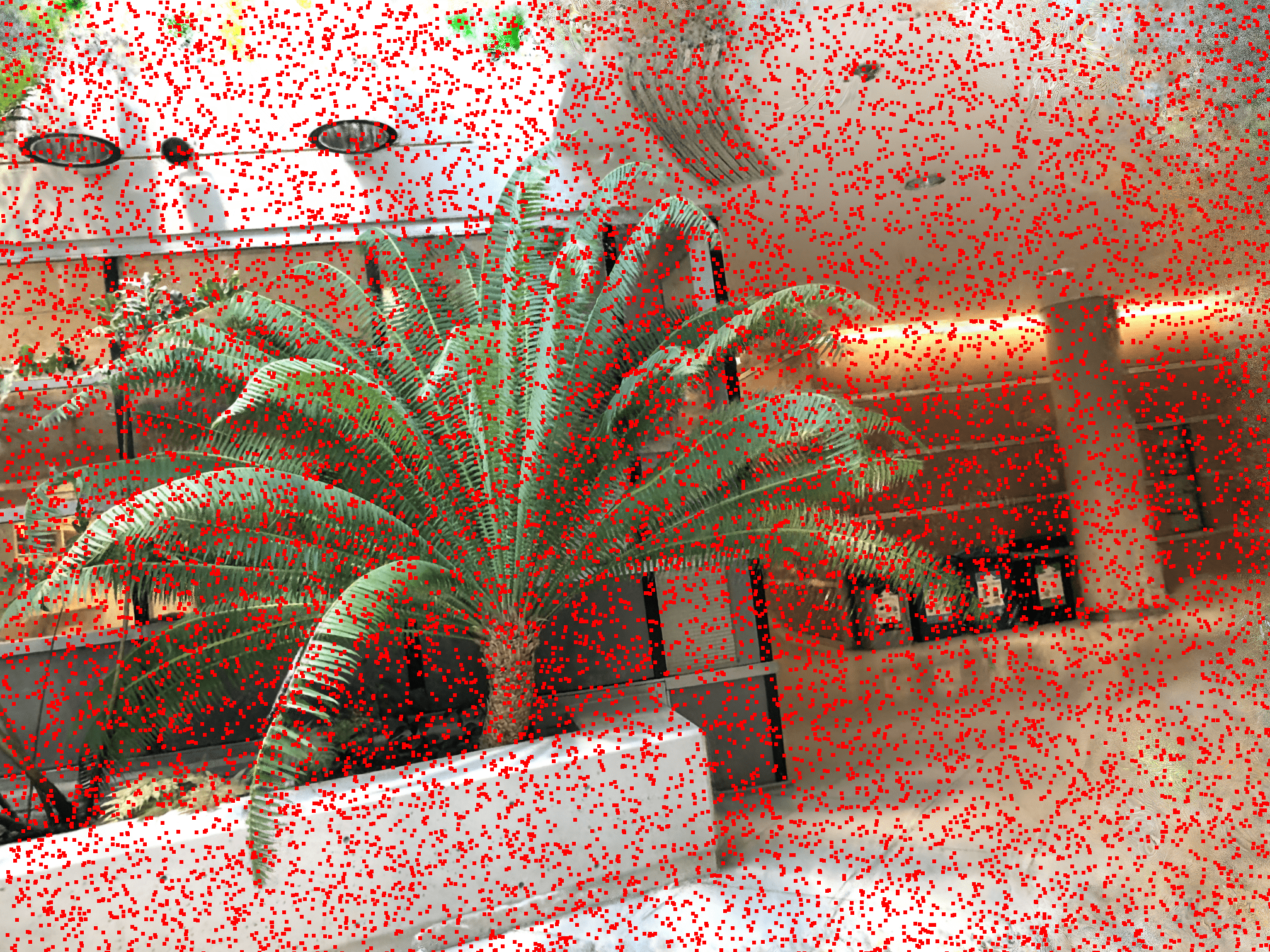}
\caption{\textbf{VF pixel sampling:} (Left) VF map of a novel view from the Fern scene from LLFF dataset (Middle) Green pixels sampled using VF map (right) Red pixels sampled randomly. The VF mask guides the process to sample pixels with more reliable RGB value.}
\label{fig:pixel_sampling}
\end{figure}
Our algorithm performs a gradient-descent optimization. Given a set of novel views, $C_A$, from scene $A$ and their corresponding viewshed maps, for each iteration $i$ sample a set of $n$ rays $R_i$ with high VF score as shown in figure \ref{fig:pixel_sampling}. Then utilize $R_i$ and the fixed NeRF parameters of scene $B$ (i.e. $\Theta_b$) to optimize the 6-DoF parameters. $\hat{T_i}$ is the estimated transformation at step $i$.
\begin{equation}\label{L_photometric}
\mathcal{L}_{\mbox{photo}}(\hat{T}_{i-1}|R_i,\Theta_b) = \frac{1}{n} \sum_{r \in R_i} \left( I(r) - I(\hat{T}_{i-1}(r)) \right)^2.
\end{equation}
The term $I(r)$ and $I(\hat{T}_{i}(r))$ denotes the pixel value generated from the ray $r$ before and after the transformation $\hat{T}_{i}$ applied.

We follow iNeRF \cite{yen2020inerf} Gradient-Based SE(3) Optimization with the following changes. First, the optimization is done over all images in the set $C_A$ instead of a single image. In other words, the rays in each batch are sampled from images across the set. This approach makes it easier to cover the scene geometry and it improves the results as can be seen in Table~\ref{Tab:ablation} (a).
Second, we use the thresholded viewshed images to select only rays with high confidence. It is important to sample only where the RGB values are known with high probability, since NeRF might have artifacts outside the region of interest. Table~\ref{Tab:ablation} (a)  shows the importance of this masking approach.

We use SGD optimizer as we found it more reliable when $T_0$ is not precise. At the end of this optimization, we get the $T \in SE(3)$ that minimizes the photometric loss (Eq.~\ref{L_photometric}). This is the rotation and translation that represent the registration between scene $A$ and scene $B$ according to our approach.
\section{Experiments}

\begin{table}
\caption{
\textbf{LLFF Results:} Rotation and translation RMS error comparison on LLFF dataset, divided into the three experiment types regarding overlap between the frames. The results are the mean error over all the scenes in the dataset, $\Delta t$ denotes the RMS translation errors multiplied by $1e2$. FPFH + RANSAC uses point clouds generated by Viewshed Fields as suggested in section \ref{initialization}. VF-NeRF + PC Init refers to our algorithm after initializing with VF-based points clouds using FPFH + RANSAC. VF-NeRF + Photo Init refers to our method using photometric based initialization.
}

\centering

\begin{tabular}{l | c c | c c | c c }

\toprule
\multirow{2}{*}{\textbf{Model}}

&\multicolumn{2}{|c|}{\textbf{Full Overlap}} 
&\multicolumn{2}{c|}{\textbf{Partial Overlap}}
&\multicolumn{2}{c}{\textbf{No Overlap}}
\\

&\textbf{$\Delta t \downarrow$} &\textbf{$\Delta R \downarrow$} 
&\textbf{$\Delta t \downarrow$} &\textbf{$\Delta R \downarrow$}
&\textbf{$\Delta t \downarrow$} &\textbf{$\Delta R \downarrow$}
\\
\midrule

FPFH~\cite{5152473} + RANSAC~\cite{10.1145/358669.358692} & 2.1149 & 1.5316 & 3.6646 & 2.7126 & 2.2679 & 2.8128
\\

iNeRF \cite{yen2020inerf}
&22.8907 & 16.1153 & 20.3248 & 12.5299 & 8.9063 & 10.0933\\

iNeRF \cite{yen2020inerf} + Photo Init.
&0.2254 & 0.1624 & 13.0270 & 2.5051 & 2.8350 & 5.6513\\

VF-NeRF + Photo Init.
& \textbf{0.0151} & 0.0206 &0.0393 & 0.0358 & 0.0324 & 0.0358\\

VF-NeRF + PC Init. & 0.0162 & \textbf{0.0157} & \textbf{0.0357} & \textbf{0.0345} & \textbf{0.0249} & \textbf{0.0286}
 \\
\bottomrule









 

\end{tabular}

\label{Tab:LLFF_results}
\end{table}


We evaluate our method on three different datasets: the standard LLFF~\cite{mildenhall2019llff} dataset, which consists of forward-facing camera poses across a plane. A small dataset of two casually collected videos captured by us that focus on object-centric scenes, and finally, the recently introduced synthetic Objaverse~\cite{deitke2022objaverse} dataset. See supplementary material for additional results and experiments.

\subsection{Real World Datasets}

\paragraph{\bf Setting:}

In the case of the LLFF and casual datasets, we run COLMAP~\cite{schoenberger2016sfm} on the entire set of images for each scene. Then, we split the frames into two sets that serve as the input to NeRF A and NeRF B, respectively. In particular, we evaluate three levels of overlap between the two sets of images. The simplest scenario is "full overlap", where NeRF A gets the even frames and NeRF B gets the odd frames. Another scenario is "partial overlap", where NeRF A gets the first 70\% even frames and NeRF B gets the last 70\% odd frames, resulting in a 40\% overlap. The last scenario is "No overlap", where NeRF A gets the first half of frames and NeRF B gets the second half of frames. 

To set the ground truth, we draw, for each experiment, a transformation $T \in SE(3)$ by randomizing the 6-DoF parameters, three rotation parameters within $[0, 45^\circ]$, and three translation parameters within $[-0.25, 0.25]$. Next we apply transformation $T^{-1}$ to the camera poses of the set of images of A, and then train VF-NeRF for both sets of images. Our goal is to find the transformation $\hat{T} \in SE(3)$ that is the closest to $T$ in terms of rotation and translation.
It should be noted that COLMAP minimizes a {\em geometric} error while we (as well as iNeRF for that matter) minimize a {\em photometric} loss. Ideally, the two should coincide but this is not always the case. We extensively discuss this in the supplemental.

We evaluated the VF-based point cloud experiments once, but due to the stochastic nature of random initialization, we evaluated all other experiments 10 times and chose the result with the highest PSNR. 
\paragraph{\bf LLFF Dataset:} Local Light Field Fusion (LLFF) \cite{mildenhall2019llff} is a widely used dataset that includes real-world complex scenes. Specifically, we evaluated VF-NeRF on the 4 common scenes - fern, trex, horns, and room. Each scene is captured from a single plane and consists of 20-62 images, depending on the scene. NeRF is very sensitive to the view direction so generating novel views in this dataset must be as near as possible to the original captured view plane.

Our results on the LLFF dataset are reported in Table~\ref{Tab:LLFF_results}. It compares a variety of methods. FPFH~\cite{5152473} + RANSAC~\cite{10.1145/358669.358692} is a point cloud based registration method. The point clouds themseleves are generated using Viewshed Fields. iNeRF~\cite{yen2020inerf}, without and with initialization, minimizes a photometric loss, like we do. Then, we evaluate our VF-NeRF with photometric based, as well as point-cloud based initializations.
\begin{table}[t]
    
\caption{
\textbf{Casually Captured Results:} Rotation and translation RMS error comparison on two Casually Captured Real-World scenes with partial overlap between the frames. $\Delta t$ denotes the RMS translation errors multiplied by $1e2$. FPFH + RANSAC uses point clouds generated by Viewshed Fields as suggested in section \ref{initialization}. VF-NeRF + PC Init refers to our algorithm after initializing with VF-based points clouds using FPFH + RANSAC. VF-NeRF + Photo Init refers to our method using photometric based initialization.
}

\centering

\begin{tabular}{l | c c | c c | c c }

\toprule
\multirow{2}{*}{\textbf{Model}}

&\multicolumn{2}{|c|}{\textbf{Lion}}
&\multicolumn{2}{c|}{\textbf{Table}}
&\multicolumn{2}{c}{\textbf{Average}}
\\

&\textbf{$\Delta t \downarrow$} &\textbf{$\Delta R \downarrow$}
&\textbf{$\Delta t \downarrow$} &\textbf{$\Delta R \downarrow$}
&\textbf{$\Delta t \downarrow$} &\textbf{$\Delta R \downarrow$}
\\
\midrule

FPFH~\cite{5152473} + RANSAC~\cite{10.1145/358669.358692} & 2.9714 & 2.9534 & 1.0842 & 1.2480 & 2.0278 & 2.1007	
\\

iNeRF \cite{yen2020inerf}
&55.3138 &17.3639 &8.7851 &16.4886 &32.0495	&16.9262\\

iNeRF \cite{yen2020inerf} + Photo Init.
&0.0639 &\textbf{0.0238} &0.0382 &0.0591 &0.0510 &0.0414\\

VF-NeRF + Photo Init. & \textbf{0.0157} & 0.0384 &0.0195 & 0.0317 
&\textbf{0.0176} &0.0351\\

VF-NeRF + PC Init. & 0.0292 & 0.0380 & \textbf{0.0125} & \textbf{0.0227} & 0.0209 & \textbf{0.0303}		\\
 
\bottomrule





\end{tabular}

\label{Tab:real_world_results}
\end{table}


As can be seen, point-cloud registration FPFH+RANSAC (first row in the table) converges but is not very accurate.
iNeRF by itself does not converge (second row of the table). It does converge in case our VF-based photometric initialization is being used (third row of the table). Finally, our method VF-NeRF converges to the most accurate solution with both types of initialization (photometric or point-cloud based). Errors below $0.05$ are hardly noticeable visually. In particular, the last row of the table shows that VF-NeRF improves the initial guess provided by FPFH+RANSAC (i.e., first row of the table) by two orders of magnitude. VF-NeRF dominates all other methods in all overlap scenarios.

\begin{table*}[!h]
   \caption{{\bf Objaverse Results:} Quantitative results of registration, organized by mean value of the best relative rotation errors $\Delta \mathbf{R}$. For example, the column titled $50\%$ we sort all scenes in ascending order of their registration rotation error and compute the mean error of the best $50\%$ scenes. $\Delta \mathbf{t}$ denotes the relative translation errors multiplied by $1e2$ with unknown scales. $\text{DReg}_{\text{df}}$ refers to DReg with density fields and DReg refers to DReg with surface field. FPFH + RANSAC uses point clouds generated by Viewshed Fields as suggested in section \ref{initialization}. VF-NeRF + PC Init refers to our algorithm after initializing with VF-based points clouds using FPFH + RANSAC. The results of FGR, REGTR and DReg are taken from ~\cite{chen2023dregnerf}.} 
  \centering
  \resizebox{0.98\textwidth}{!}{
    \begin{tabular}{l | c c | c c | c c | c c }
      \toprule
      \multirow{2}{*}{\textbf{Model}}

        &
        \multicolumn{2}{c|}{\textbf{50\%}} & \multicolumn{2}{c|}{\textbf{75\%}} &  
        \multicolumn{2}{c|}{\textbf{90\%}} & \multicolumn{2}{c}{\textbf{100\%}}
      \\
        & $\Delta \mathbf{t}$ & $\Delta \mathbf{R}$ & 
        $\Delta \mathbf{t}$ & $\Delta \mathbf{R}$ & 
        $\Delta \mathbf{t}$ & $\Delta \mathbf{R}$ & 
        $\Delta \mathbf{t}$ & $\Delta \mathbf{R}$
        \\
      
      \midrule

      FGR~\cite{zhou2016fast} & 5.33 & 13.20 & 8.23 & 16.98 & 12.79 & 46.85 & 13.90 & 61.59
                                           \\
      REGTR~\cite{yew2022regtr} & 35.08 & 65.98 & 42.87 & 93.84 & 43.11 & 105.58 & 43.31 & 113.78
                                           \\
      $\text{DReg}_{\text{df}}$~\cite{chen2023dregnerf} & 5.43 & 18.62 & 12.17 & 56.82 & 14.45 & 74.32 & 16.06 & 86.23
      \\
      DReg~\cite{chen2023dregnerf} & 3.24 & 5.33 & 3.61 & 7.38 & 3.77 & 8.59 & 3.85 & 9.65
                         \\
      FPFH~\cite{5152473} + RANSAC~\cite{10.1145/358669.358692} & 1.94 & 1.96 & 1.82 & 2.79 & 2.19 & 3.69 & 3.01 & 9.75
                         \\
      VF-NeRF + PC Init.    & \textbf{0.47} & \textbf{0.26} & \textbf{0.57} & \textbf{0.60} & \textbf{1.08} & \textbf{1.11} & \textbf{2.14} & \textbf{6.77}
      \\

      \bottomrule
    \end{tabular}
}

   \label{table:quantitive_objaverse_registration}
\end{table*}

\paragraph{\bf Casually Captured Scenes}
One of the advantages of NeRF is the simplicity of capturing a scene in-the-wild and reconstructing the 3D model directly from the video. To evaluate the capabilities of NeRF registration on such scenes, we captured two $360^\circ$ scenes using a camera phone. Each scene includes 300-400 images and an example of generated point clouds of the scenes can be seen in figure \ref{generated_pc}, more images from the dataset appear in the supplementary. We repeat the same experiment we did with the LLFF dataset and compare with the same methods. The results are reported in Table \ref{Tab:real_world_results}. As can be seen, results are very similar to the LLFF experiment. VF-NeRF (with either one of the possible initializations) dominates other methods.

\begin{figure}[t]
    \centering
    \begin{subfigure}[b]{0.24\linewidth}
        \includegraphics[width=\textwidth]{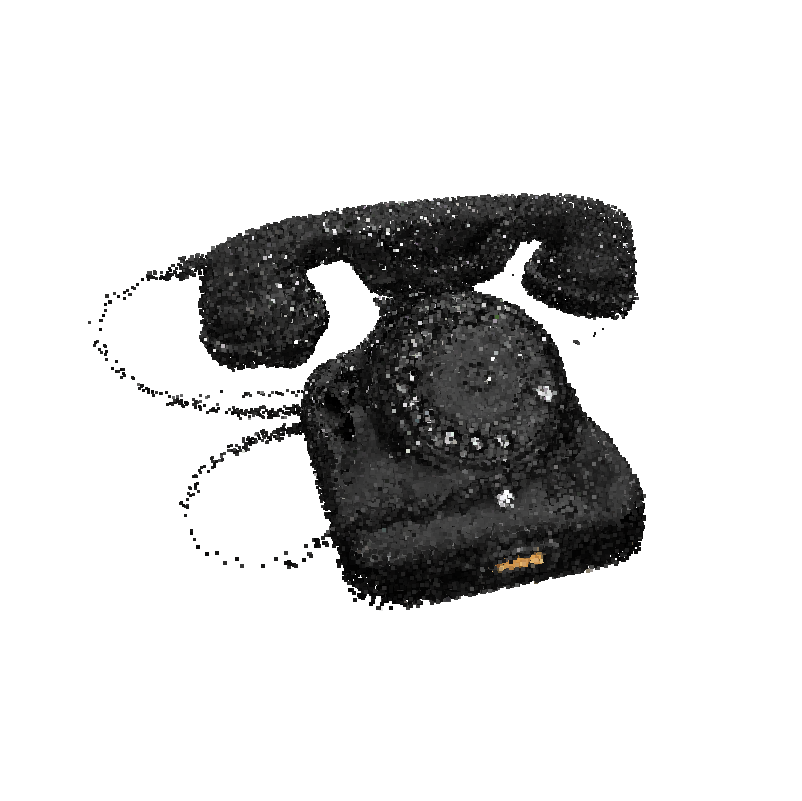}
        \caption{}
        \label{fig:pc_phone}
    \end{subfigure}
    \begin{subfigure}[b]{0.24\linewidth}
        \includegraphics[width=\textwidth]{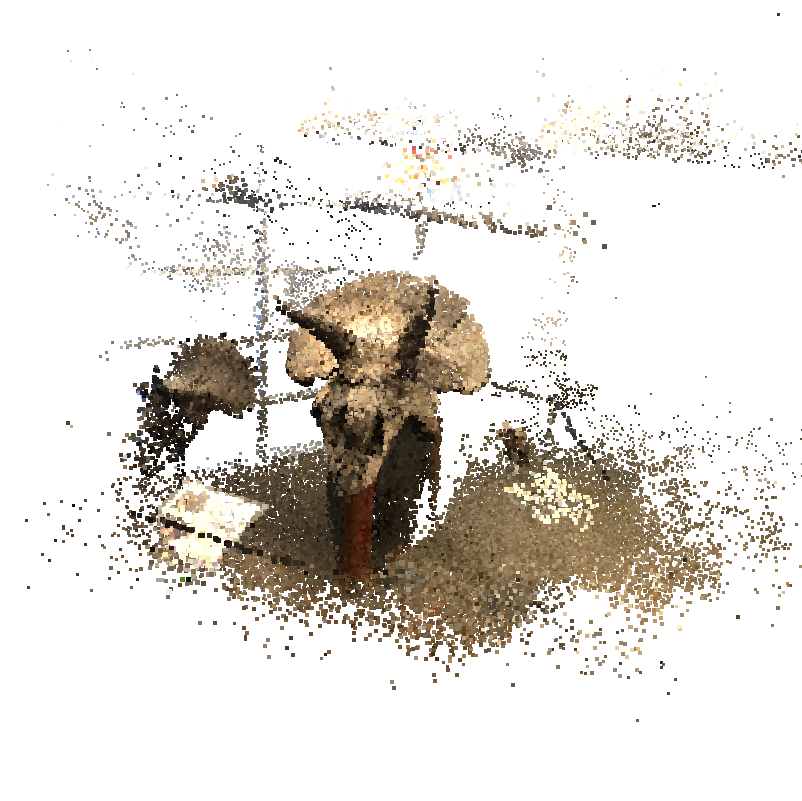}
        \caption{}
        \label{fig:pc_horns}
    \end{subfigure}
    \hfill
    \begin{subfigure}[b]{0.24\linewidth}
        \includegraphics[width=\textwidth]{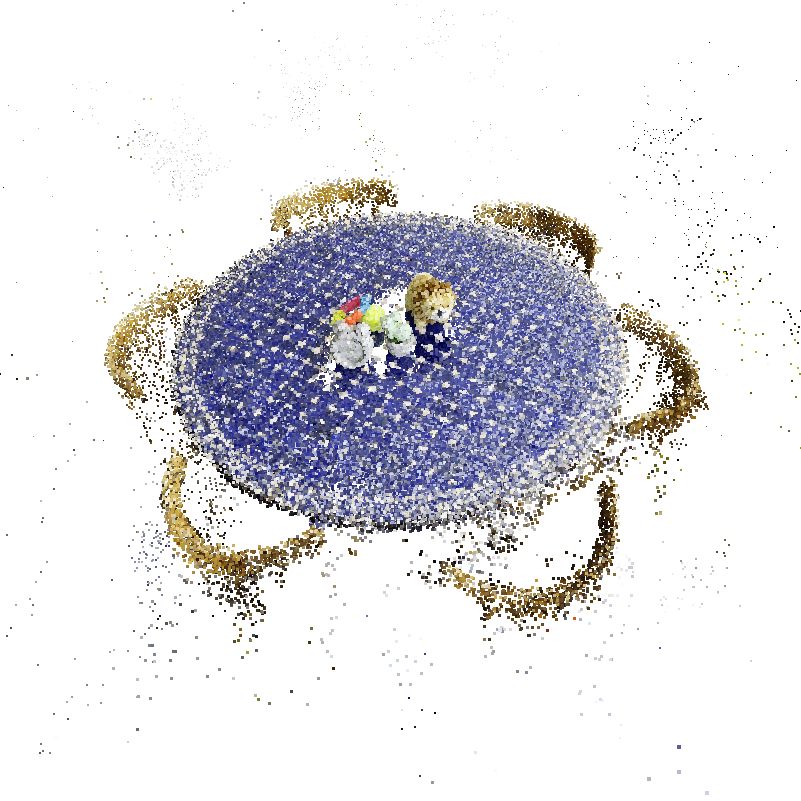}
        \caption{}
        \label{fig:pc_table}
    \end{subfigure}
    \hfill
    \begin{subfigure}[b]{0.24\linewidth}
        \includegraphics[width=\textwidth]{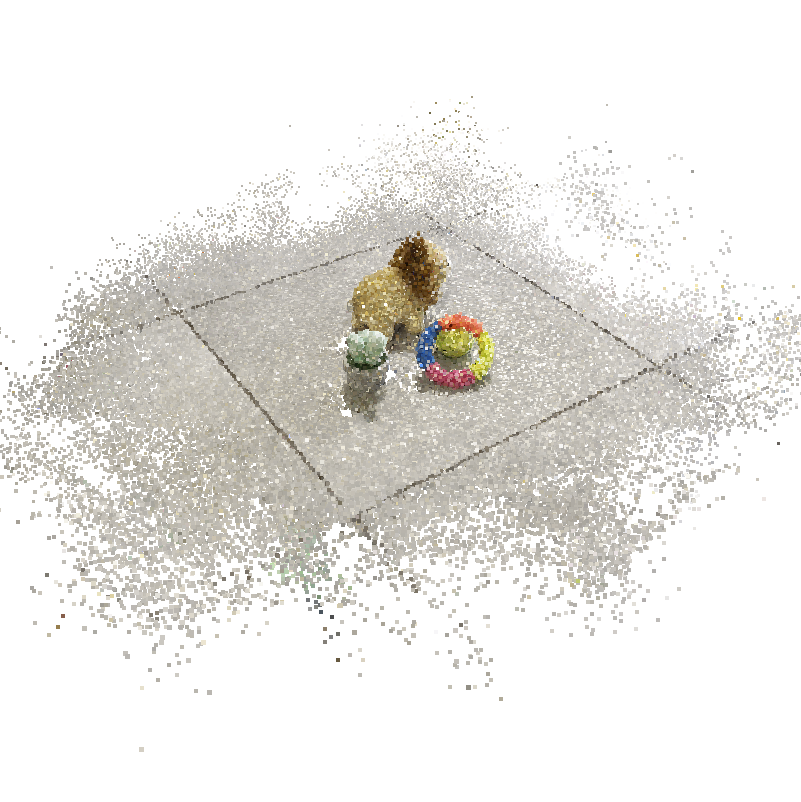}
        \caption{}
        \label{fig:pc_lion}
    \end{subfigure}
\caption{
\textbf{Point clouds from VF:} Point clouds generated by sampling from the VF distribution as explained in sub section~\ref{initialization}. Each point cloud here is a combination of two point clouds from two NeRFs after applying our registration method. The examples are taken from all the datasets we evaluate in the paper, \ref{fig:pc_phone} from Objaverse dataset, \ref{fig:pc_horns} from LLFF dataset and \ref{fig:pc_table}-\ref{fig:pc_lion} from our casually captured dataset.
} 
\label{generated_pc}
\end{figure}

\subsection{Objaverse Dataset}
Objaverse\cite{deitke2022objaverse} is a large dataset that contains more than 800K 3D objects. DReg~\cite{chen2023dregnerf} utilized this dataset to construct a dataset of 1700+ training scenes and 44 evaluation scenes for the task of NeRF registration. Since our method is optimization based, the training set is not required and we evaluated our method directly on the 44 evaluation scenes. To initialize our method we first used VF to sample oriented points, choose points with density larger than 10 and reconstructed a point cloud for each scene. Then we used FPFH~\cite{5152473} + RANSAC~\cite{10.1145/358669.358692} to find an initial alignment between the two point clouds.

We compare VF-NeRF to several point-cloud based registration methods, including FGR~\cite{zhou2016fast}, REGTR~\cite{yew2022regtr}, FPFH~\cite{5152473}+RANSAC~\cite{10.1145/358669.358692}, and two flavours of DREG~\cite{chen2023dregnerf}. We report results in  Table~\ref{table:quantitive_objaverse_registration}. The table consists of multiple columns, that correspond to different fractions of the dataset. For example, in the column titled $50\%$ we sorted all 44 scenes in ascending order, according to their rotation error, took the 22 scenes with the lowest error and computed their mean error. As can be seen, we dominate the table on both rotation and translation errors. Specifically, our method shows a notable improvement over FPFH~\cite{5152473} + RANSAC~\cite{10.1145/358669.358692} (that serves as initialization to our VF-NeRF method).

We conclude that VF-NeRF performs well across all types of scenes both real and synthetic, and that VF can be used to help initialize the optimization.

\begin{table}
\caption{\textbf{(a) Ablation study:} impact of each component of our method on performance. We report mean RMS error of rotation and translation over all scenes in the LLFF dataset. $\Delta t$ denotes the RMS translation errors multiplied by $1e2$ \textbf{(b) Noise robustness study:} 
Impact of noise on estimation the position of the oriented points. Results are reported on the Trex scene from the LLFF dataset. The noise column denotes the amount of uniform noise (as percentage of scene size) that is added. The results are the mean error over the three overlap settings. As can be seen, VF-NeRF is robust to this noise.}
\centering
\begin{tabular}{ccc}

\begin{tabular}{lcc}

\toprule
\textbf{Model} &\textbf{$\Delta t \downarrow$} &\textbf{$\Delta R \downarrow$}\\
\midrule

No initialization 
&39.2993 &21.5474\\

No VF Masks
&1.7348 &1.3609\\

Single image
&9.4008 &8.9252\\

\textbf{Full method}
&\textbf{0.0151} &\textbf{0.0206}\\
 
\bottomrule

\end{tabular} & ~~~~~~~~~~~~~ &

\begin{tabular}{lcc}

\toprule
\textbf{Noise} &\textbf{$\Delta t \downarrow$} &\textbf{$\Delta R \downarrow$}\\
\midrule

0\%  &
\textbf{0.0113} & 0.0108\\
1\% &
0.0114 & 0.0120\\
5\% &
0.0117 & \textbf{0.0106}\\
10\% &
0.0265 & 0.0140\\
20\% &
0.0149 & 0.0155\\

\bottomrule

\end{tabular} 
\\

(a) Ablation Study & ~~~~~~~~~~~~~ & (b) Noise robustness study\\
\end{tabular}

\label{Tab:ablation}
\end{table}

\subsection{Ablation Study}

We conducted a number of ablation studies to evaluate different aspects of our approach.

\paragraph{\bf VF Abalation:}

Table~\ref{Tab:ablation} (a) shows the contribution of each component of our method to the overall solution. The experiment was conducted on the LLFF dataset using the exact same settings reported for Table~\ref{Tab:LLFF_results}.

As can be seen, changing even a single component of the full method causes dramatic performance degradation. Not surprisingly, the most significant degradation happened when we disabled the initialization technique to find $T_0$. NeRF tends to be accurate in the region of interest, but produces completely meaningless results outside it. Hence, without initialization the photometric loss in this case is useless.      
Another major performance degradation occurred when no viewshed maps were used. This can be be explained by looking at our viewshed-based sampling example, shown in Figure~\ref{fig:pixel_sampling}. When generating novel views of unknown NeRF scenes it is very likely to capture some areas out of the original region of interest, which is basically noise in terms of NeRF.
\begin{figure}
    \centering
    \begin{minipage}[t]{0.49\textwidth}
        \vspace{0pt}
        \begin{subfigure}[b]{0.48\linewidth}
            \includegraphics[width=\textwidth]{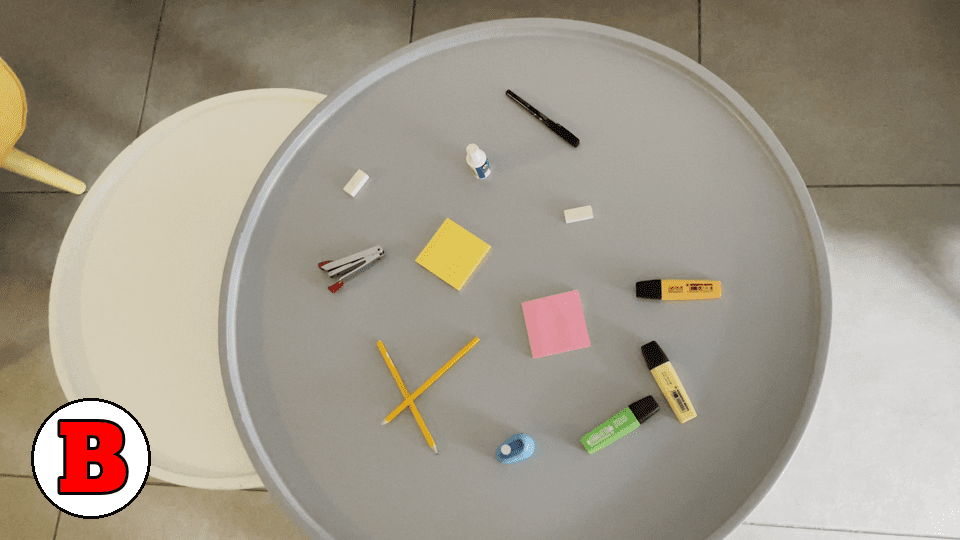}
        \end{subfigure}
        \begin{subfigure}[b]{0.48\linewidth}
            \includegraphics[width=\textwidth]{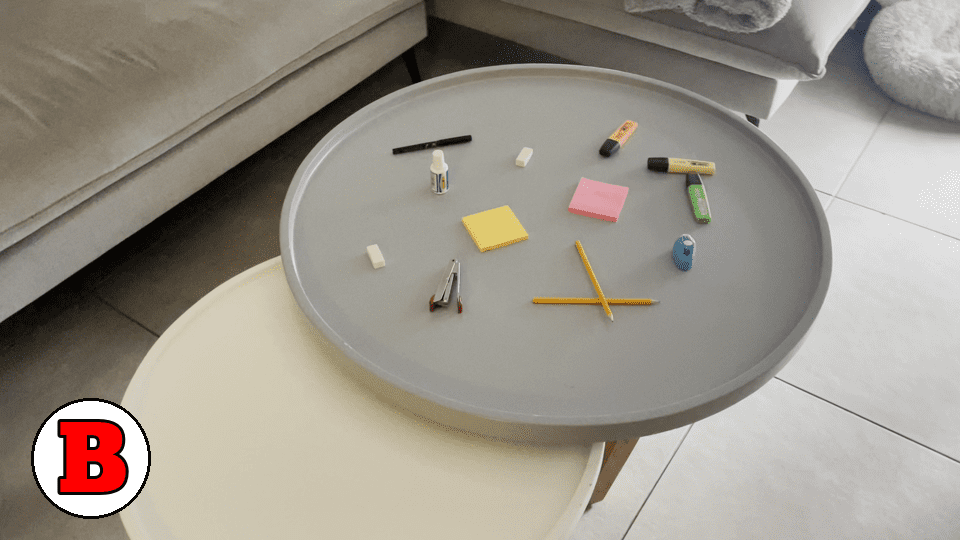}
        \end{subfigure}
        \begin{subfigure}[b]{0.48\linewidth}
            \vspace{2pt}
            \includegraphics[width=\textwidth]{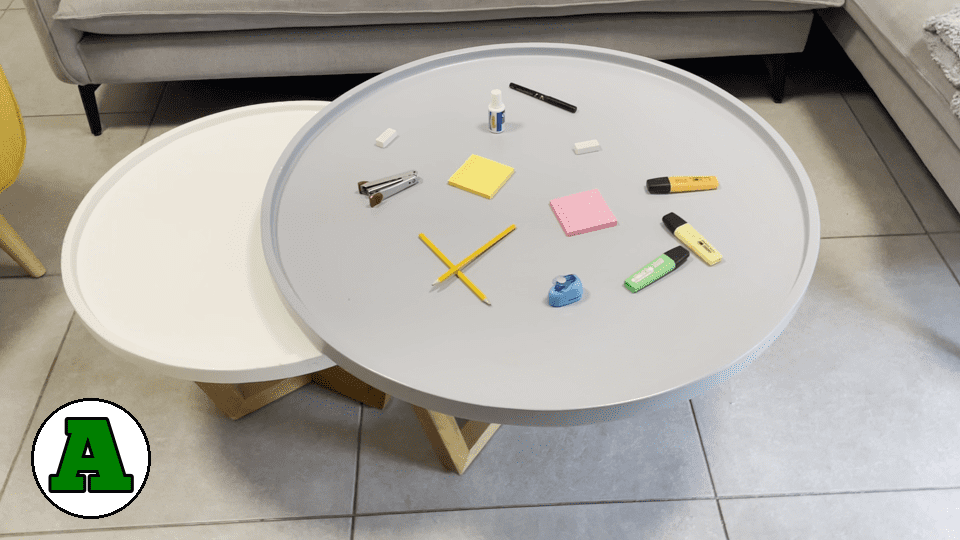}
        \end{subfigure}
        \hspace{0pt}    
        \begin{subfigure}[b]{0.48\linewidth}
            \vspace{2pt}
            \includegraphics[width=\textwidth]{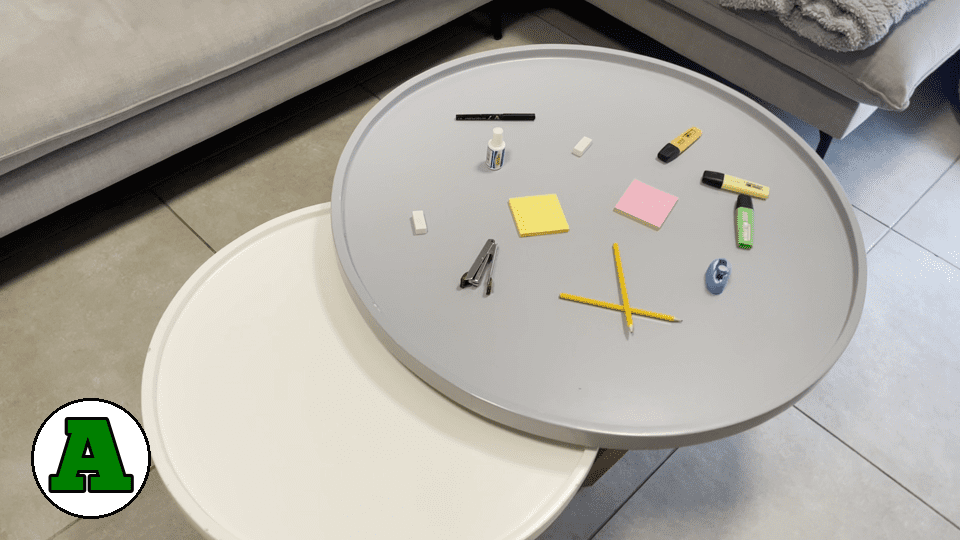}
        \end{subfigure}
    \end{minipage}
    \hfill
    \begin{minipage}[t]{0.49\textwidth}
    \vspace{0pt}
    \begin{subfigure}[b]{0.99\linewidth}
        \includegraphics[width=\textwidth]{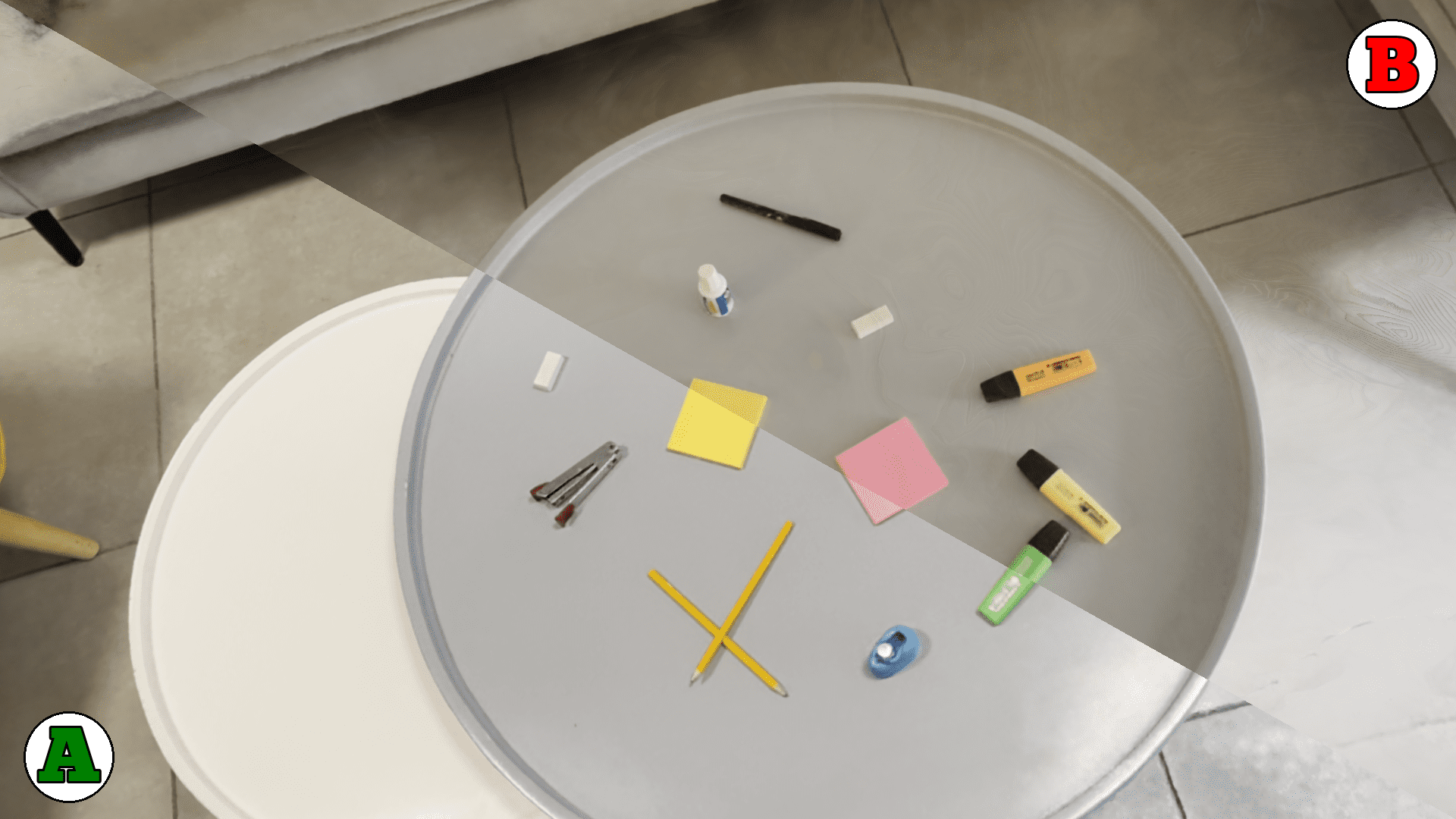}
    \end{subfigure}
    \end{minipage}
\caption{
\textbf{Illumination:} (Left) 4 images from the original videos, top images taken from one video and the bottom images from another video. Each video was taken under different illumination condition. (Right) Registration of the two scenes where the bottom left part is taken from one NeRF and the top right part is taken from another NeRF.} 
\label{illumination}
\end{figure}
The viewshed maps filter out the noisy pixels and thus the photometric loss (Equation~\ref{L_photometric}) is applied to more reliable pixels.

\paragraph{\bf Noisy VF:} In the next ablation study we add noise to the oriented points used to construct VF. Specifically, we add up to $20\%$ (of the size of the scene) zero-mean noise to the position of the oriented points that are used to train the NF algorithm. Then, we generated camera viewpoints by sampling the NF latent space, as before. Table~\ref{Tab:ablation} (b) shows the results. It is interesting to note that even when noise of up to $20\%$ of the size of scene is added to the position of the oriented points, performance do not degrade. We suspect this is because the goal of the oriented points, that are fed to the Normalizing Flows algorithm, is simply to generate a plausible {\em initialization} for the position of the virtual cameras. This is enough to produce sufficiently good novel images for the optimization.

\paragraph{\bf Illumination changes}
Our method minimizes a photometric loss, it is therefor a fair question to ask if it works with different illumination conditions? To test that we captured a scene twice, each time with different illumination, and then reconstructed a NeRF for each instance. We ran COLMAP on each video separately to extract the pose estimation, and found that this has a several consequences. First, each NeRF has its own coordinate system so there is no ground-truth and the results are qualitative. Second, each NeRF has its own arbitrary scale factor since COLMAP only estimates extrinsics and intrinsics up to scale. In order to deal with this issue, we added a learnable scale factor to our method on top of the 6-DoF learnable parameters. This is a generalization of the base method for the case of scenes with unknown scale. Our qualitative results are shown in Figure~\ref{illumination}. On the left we show two frames from each of the videos. Observe how the color of the table changes, as a result of illumination change. On the right we show the scene from a novel view point where images from both NeRFs are registered using VF-NeRF. As can be seen, the registration is quite accurate, despite the change in illumination. This is especially important because VF-NeRF minimizes a photometric loss. 

\begin{figure}[t]
\centering
\includegraphics[width=0.5\textwidth]{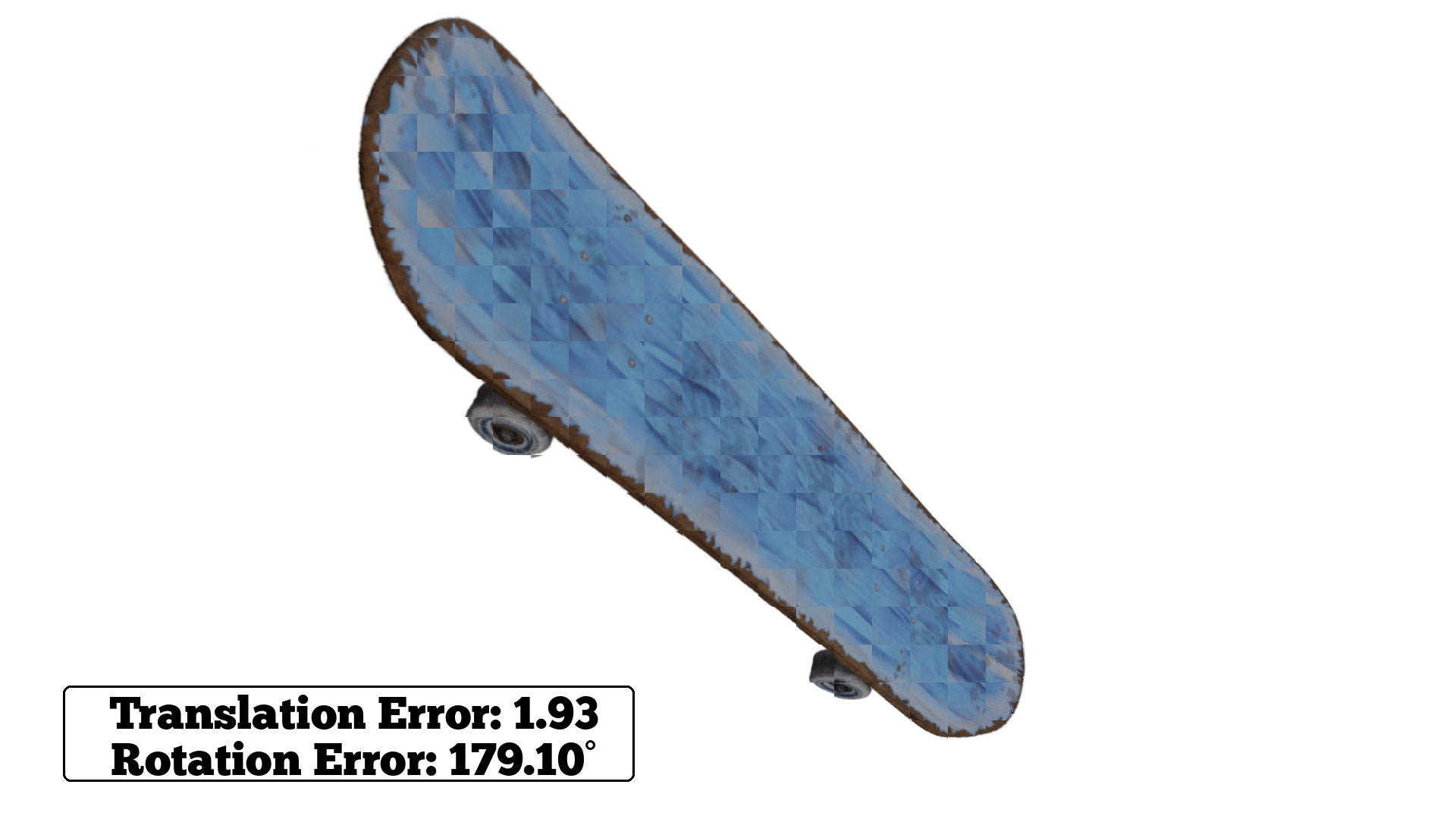}
\hfill
\caption{\textbf{Failure Example:} Failure example of a scene from Objaverse dataset that converged to a wrong solution. The registered scene is upside-down with respect to the original scene. The image is from a novel view point and is created from the two registered NeRFs where we alternate between them using a checkerboard pattern of 50 pixels (zoom in to see the pattern artifacts on the skateboard).}
\label{fig:Failure_example}
\end{figure}

\paragraph{\bf Limitations}
VF-NeRF registration relies on photometric loss, a method susceptible to inaccuracies when applied to textureless surfaces. Moreover, VF-NeRF may converge to a partially accurate solution in terms of photometric loss. however, in scenes exhibiting high symmetry, this solution may deviate by $180^\circ$ from the true solution, as illustrated in Figure \ref{fig:Failure_example}. It would also be interesting to stress test VF-NeRF on scenes with partial overlap, or scenes with moving objects.

\paragraph{\bf Why not use COLMAP?} It is certainly possible to use COLMAP from scratch to register images from both NeRFs. But, we believe that NeRF registration can provide an efficient building block for modular and scaleable pipeline, where small groups of images are bundle adjusted together into small NeRFs that are later registered into larger NeRF models. In addition, Viewshed Fields, are quite a useful tool for sampling "good" novel view points as well as sampling "good" 3D points of the scene. Going forward, there might be novel applications that will benefit from NeRF registration in general, and VF in particular.

\section{Conclusion}
We considered the problem of NeRF registration and suggested a novel representation, termed Viewshed Fields (VF), to help solve it. VF is an implicit function, just like NeRF, that captures the likelihood of 3D surface points to be viewed by the original cameras. The novel combination of VF with Normalizing Flows (NF) helps in various use cases. It can be used to sample novel camera view points, on one hand, or to sample a colored 3D point cloud, on the other. It can also be used to guide ray sampling during the optimization of NeRF registration. We evaluated our method, VF-NeRF, on several diverse datasets that include front-facing scenarios, object-centric videos, and images of synthetic objects, and achieved SOTA results on many of them.

\section{Supplementary}
\section{Appendix Overview}
This document contains supplemental material regarding the following topics:
\begin{enumerate}
    \item \textbf{Appendix Overview}
    \item \textbf{Original Images} - Discussion about the differences between using generated novel views and the original images.
    \item \textbf{Noise Robustness Study} - Visualization of the noise robustness study.
    \item \textbf{VF Direction} - Discussion and visualization of the importance of direction $\Vec{d}$ to the VF
    \item \textbf{Registration Examples} - Videos of VF-NeRF registration results of different scenes.
    \item \textbf{"Casually Captured" Dataset} - Examples from our casually captures scenes.
    \item \textbf{Implementation Details} - Full detailed implementation details including computation, parameters etc. 
    \item \textbf{Full Results} - The results of all the experiments, represented scene by scene.
\end{enumerate}
\begin{figure}
    \centering
    \begin{subfigure}[b]{0.32\linewidth}
        \includegraphics[width=\textwidth]{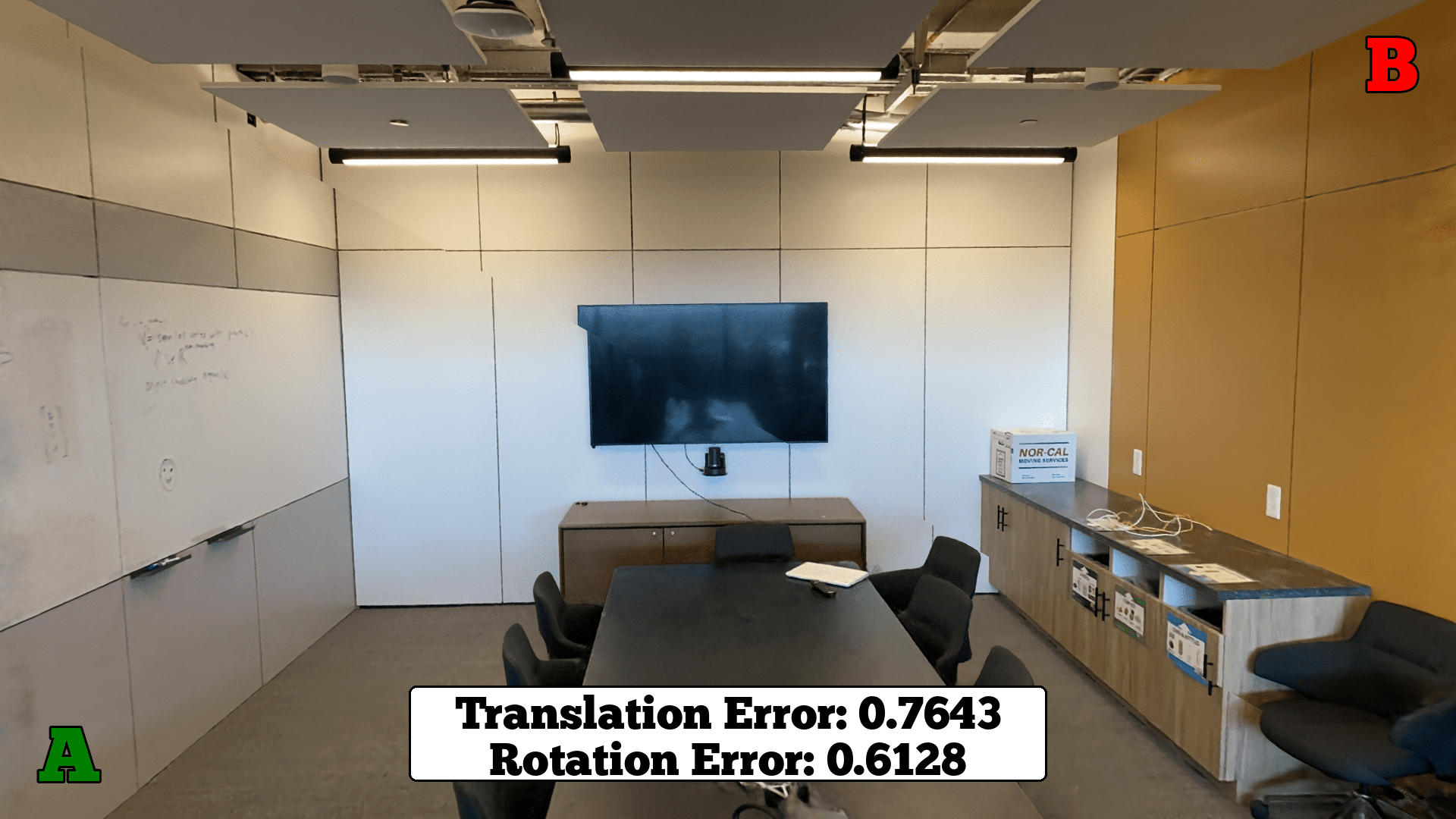}
        \caption{}
        \label{fig:room_merged_inerf}
    \end{subfigure}
    \hfill
    \begin{subfigure}[b]{0.32\linewidth}
        \includegraphics[width=\textwidth]{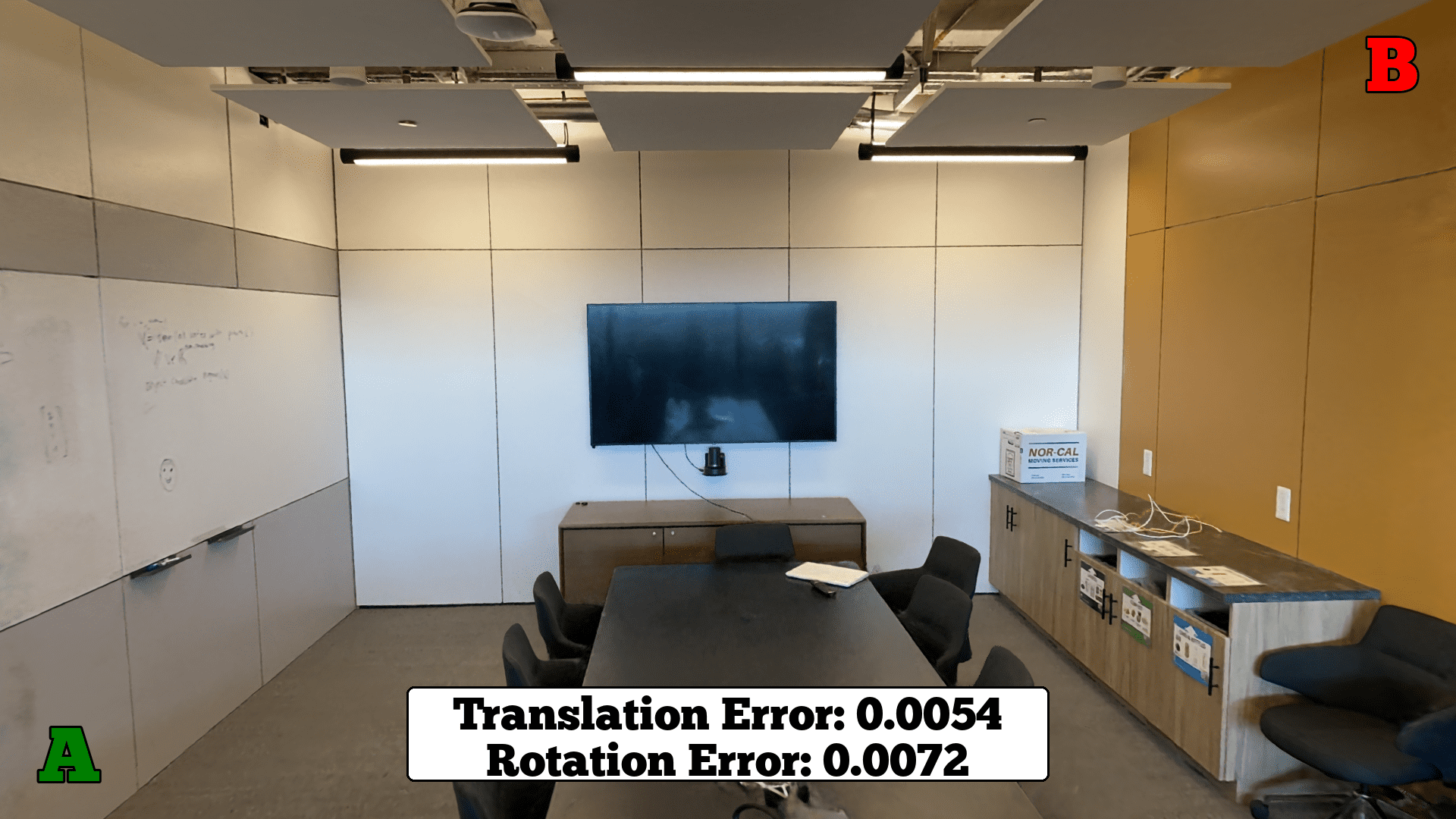}
        \caption{} 
        \label{fig:room_merged_ours}
    \end{subfigure}
    \hfill
    \begin{subfigure}[b]{0.32\linewidth}
        \includegraphics[width=\textwidth]{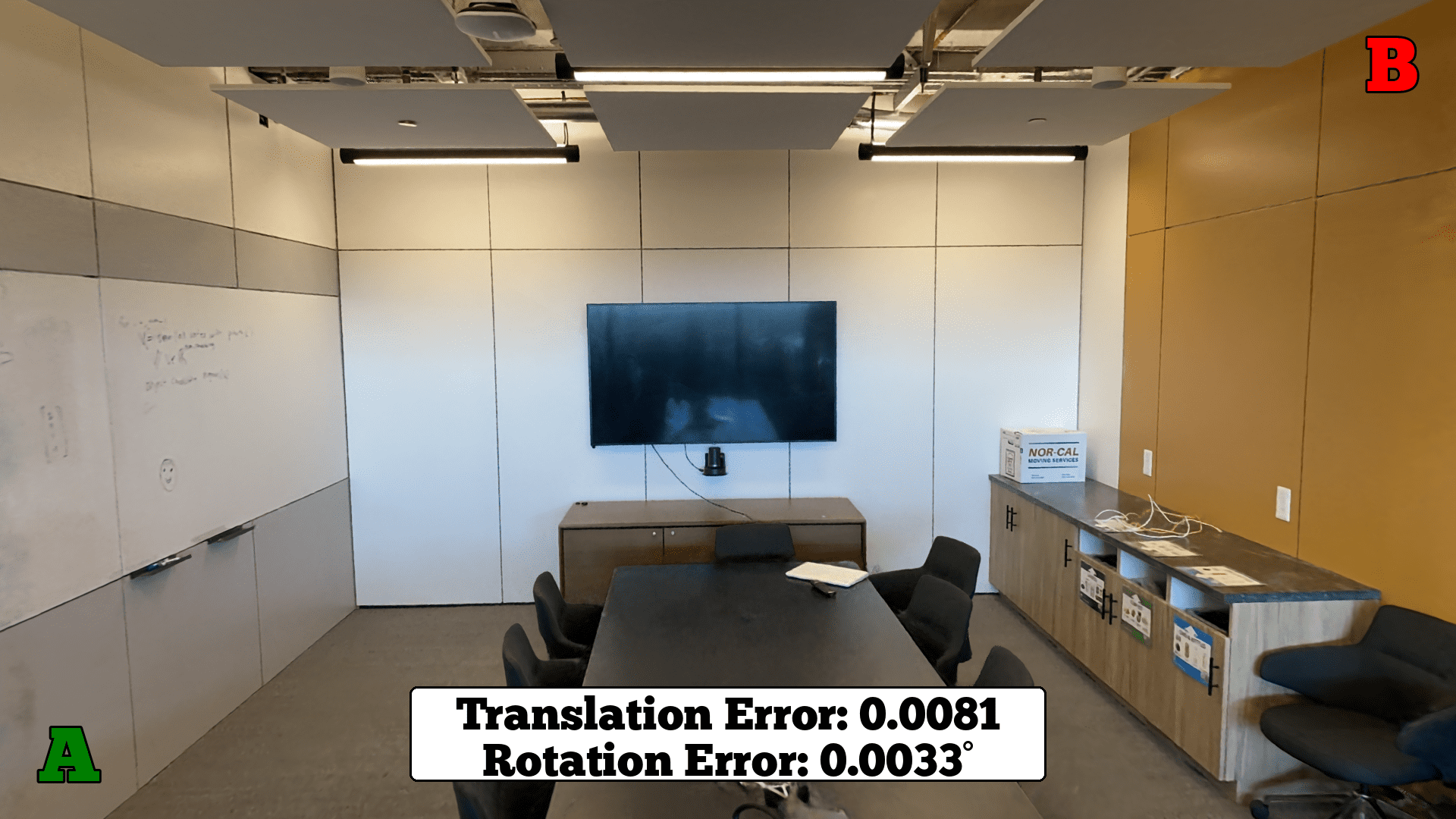}
        \caption{} 
        \label{fig:room_merged_orig_cams}
    \end{subfigure}
\caption{
\textbf{Error Visualization:} A visualization of the registration results of different methods, this visualization clarifies the visual impact behind the RMS error numbers. \ref{fig:room_merged_inerf} is INeRF based (Observe the mismatch on the edge of the TV), \ref{fig:room_merged_ours} is our method, \ref{fig:room_merged_orig_cams} is original cameras based, all three initialized with our VF. Note that the difference between \ref{fig:room_merged_ours} and \ref{fig:room_merged_orig_cams} is visually indistinguishable.} 
\label{error_visualization}
\end{figure}

\section{Original Images}
Our method is completely agnostic to the original images that were used to generate the NeRF 3D scene, but it is a fair question to ask - why not use them for the registration process instead of using NF and viewshed maps?

To address this question we evaluated an extensive experiment over the LLFF dataset \ref{Tab:original_cameras} comparing our method to registration based on the original images, the experiment settings are exactly the same as the LLFF dataset experiment.
Table~\ref{Tab:original_cameras} shows that we can get comparable results with both the original images and the generated VF images. Moreover, we argue that although the error is numerically lower when using the original images, it is visually indistinguishable as shown in figure \ref{error_visualization}. 

 An explanation for the numerical difference is that we use COLMAP as a ground truth for determining camera poses, which is subject to estimation inaccuracies too. the inaccuracies might be different between the methods since COLMAP minimizes \textbf{geometric} objective function, while our method minimizes \textbf{photometric} objective function. Thus, in scenarios with minimal registration error within a noisy setting, discerning the superiority between COLMAP's error and the photometric error remains ambiguous.
\begin{table}
\caption{
\textbf{Using Viewshed Fields Vs. Original Images:} We compare registration error when using the original images vs. using VF-NeRF. Comparison is done on the LLFF dataset in three different settings and the results are the mean error over all the scenes in the dataset. $\Delta t$ denotes the RMS translation errors multiplied by $1e2$. As can be seen, the error of both methods is under 0.05 for rotation and translation in all categories.
}
\centering

\begin{tabular}{l | c c | c c | c c }

\toprule
\textbf{Model} 
&\textbf{$\Delta t \downarrow$} &\textbf{$\Delta R \downarrow$}
&\textbf{$\Delta t \downarrow$} &\textbf{$\Delta R \downarrow$}
&\textbf{$\Delta t \downarrow$} &\textbf{$\Delta R \downarrow$}
\\
\midrule

&\multicolumn{2}{|c|}{\textbf{Full Overlap}}
&\multicolumn{2}{c|}{\textbf{Partial Overlap}}
&\multicolumn{2}{c}{\textbf{No Overlap}}
\\

\midrule

Original images + Photo init
&\textbf{0.0132} & \textbf{0.0099} & \textbf{0.0182} & \textbf{0.0089} & \textbf{0.0188} & \textbf{0.0143}\\

VF-NeRF + Photo Init.
& 0.0151 & 0.0206 &0.0393 & 0.0358 &0.0324 & 0.0358\\

\bottomrule











\end{tabular}

\label{Tab:original_cameras}
\end{table} 

\section{Noise Robustness Study}
\begin{figure}[t]
\centering
    \begin{subfigure}[b]{0.49\linewidth}
        \includegraphics[width=\textwidth]{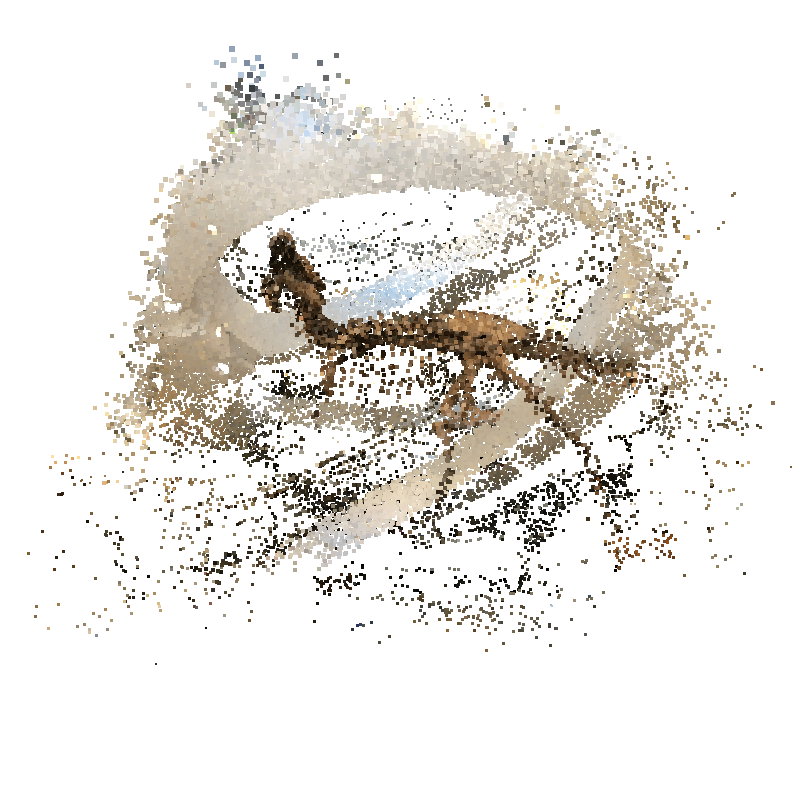}
    \end{subfigure}
    \hfill
    \begin{subfigure}[b]{0.49\linewidth}
        \includegraphics[width=\textwidth]{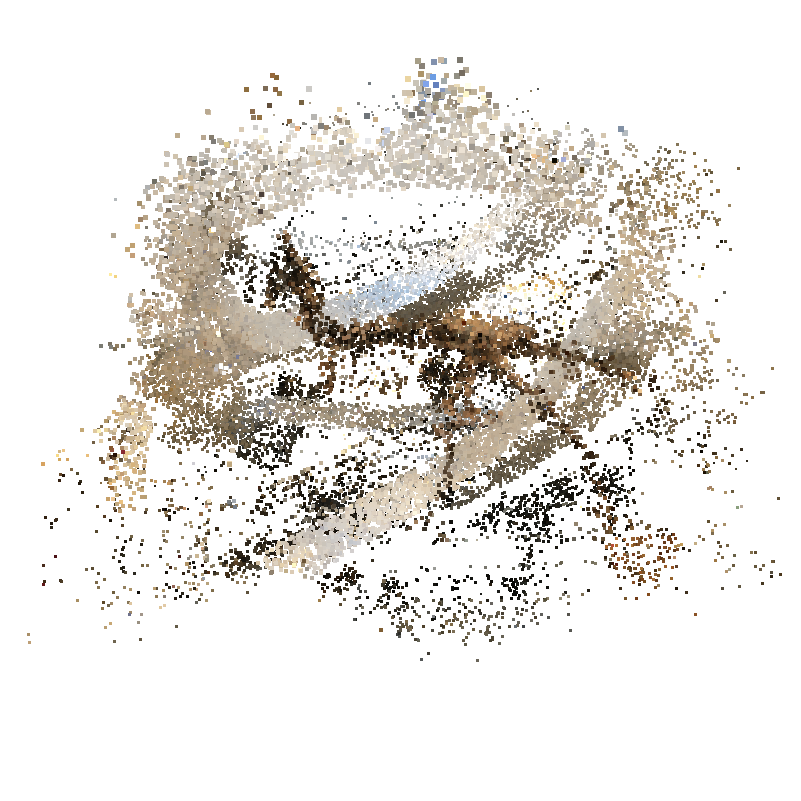}
    \end{subfigure}
    \hfill
\caption{\textbf{Noise Robustness Study Point Clouds:} Two point clouds of the Trex (LLFF) scene generated by VF samples. (Left) A Point cloud from a scene without noised oriented points. (Right) A Point cloud from a scene with noised oriented points (20\% of the scene scale with zero-mean distribution).}
\label{fig:noise_pc}
\end{figure}
Figure \ref{fig:noise_pc} illustrates the impact of introducing noise to oriented points through point cloud visualization. A noticeable decline in quality is observed when comparing the point clouds of the normal and noised scenes. However, it also reveals that the generated oriented points from the noised scene are capable of generating a reasonably accurate point cloud, thus enabling the creation of satisfactory novel views, as demonstrated in the noise robustness study section of the main paper.

\section{VF direction}
\begin{figure}[t]
\centering
    \begin{subfigure}[b]{0.99\linewidth}
        \includegraphics[width=\textwidth]{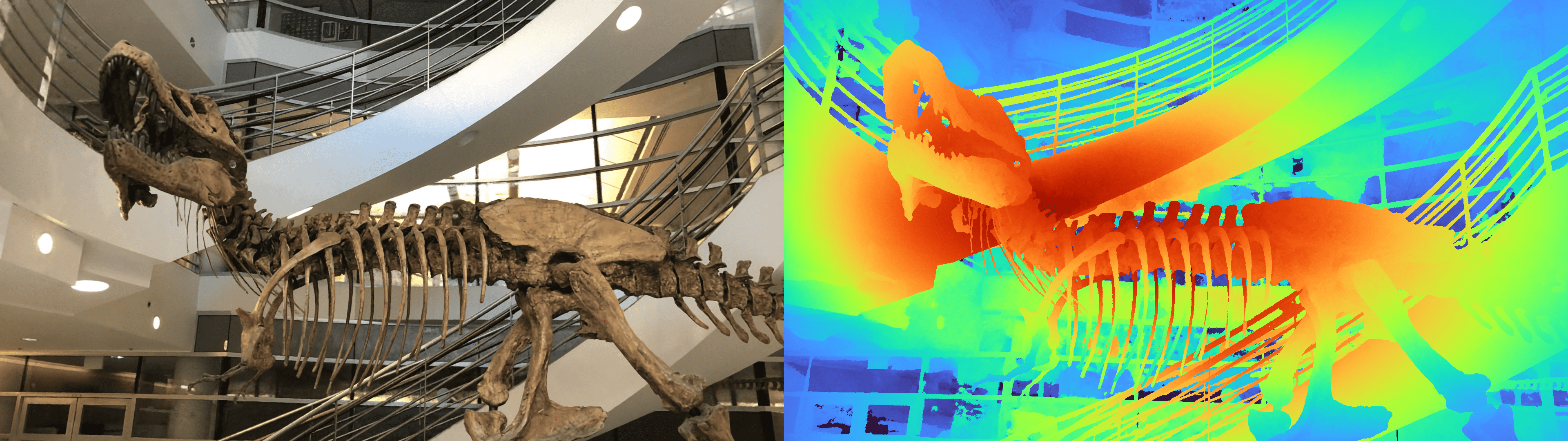}
    \end{subfigure}
    \hfill
    \begin{subfigure}[b]{0.99\linewidth}
        \includegraphics[width=\textwidth]{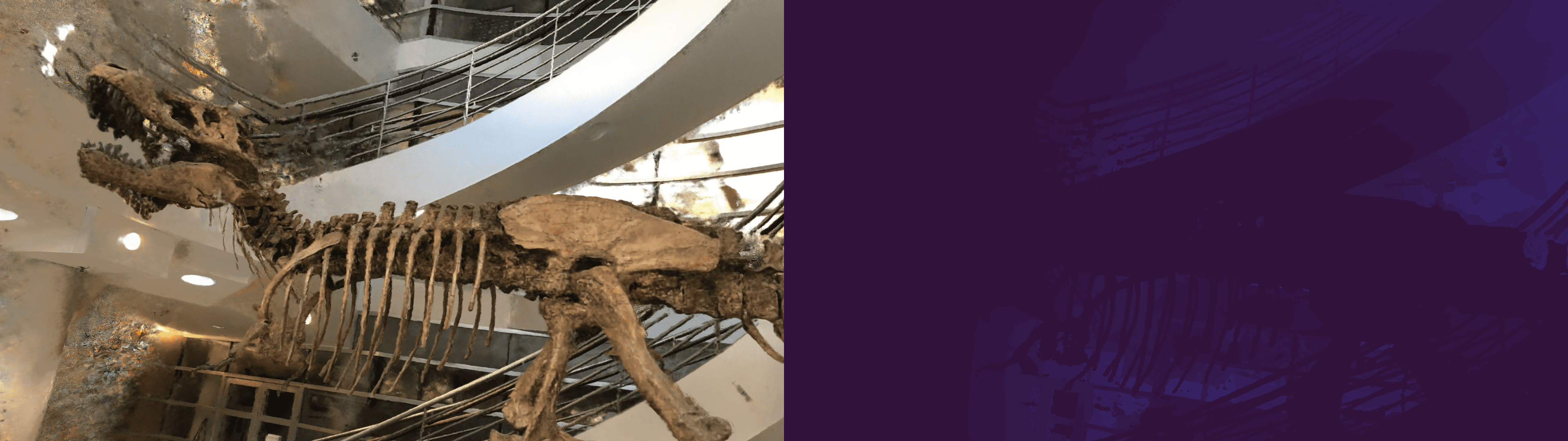}
    \end{subfigure}

\caption{\textbf{Viewshed Fields Direction:} (Top) Example of novel view and its corresponding VF map from the cameras plane of LLFF Trex scene, which is a good looking direction (Bottom) Example of novel view and its corresponding VF map from the side of LLFF Trex scene, which is a bad looking direction. Note that the oriented points are in the same location $\Vec{x}$ but different direction $\Vec{d}$ with a major affect on the image quality and VF values as expected.}
\label{fig:VF_direction}
\end{figure}
Viewshed Fields (VF) maps oriented point $(\Vec{x},\Vec{d})$ to latent space $\Vec{z}$. When looking at a generated VF map, it is clear that oriented points with high values are located (\ie $\Vec{x}$) in the ROI, and oriented points with low values are located outside the ROI. But, looking at a single image is not enough to understand the importance of the direction $\Vec{d}$. To clarify the important effect of the direction $\Vec{d}$ to the VF, we generated VF of the same object from different directions. \Cref{fig:VF_direction} demonstrates the crucial effect of direction on the VF, which supports the novel view generation by estimation of a good-looking direction. For a better understanding of the effect of direction on the VF, see the attached video.

\section{Registration Examples}
Throughout the paper, we visualize the registration results using novel views and point clouds. We use videos to show the complete scene registration, as they cover the whole scene from different viewpoints. See the attached videos for registration examples on three different scenes. Note that there are some artifacts ("floaters") in the videos, these artifacts are not related to the registration process but rather to the NeRF quality of some novel views, as expected when one of the scenes is not originally well covered from the novel viewpoint.

\section{"Casually Captured" Dataset}
\begin{figure}[t]
\centering
    \begin{subfigure}[b]{0.24\linewidth}
        \includegraphics[width=\textwidth]{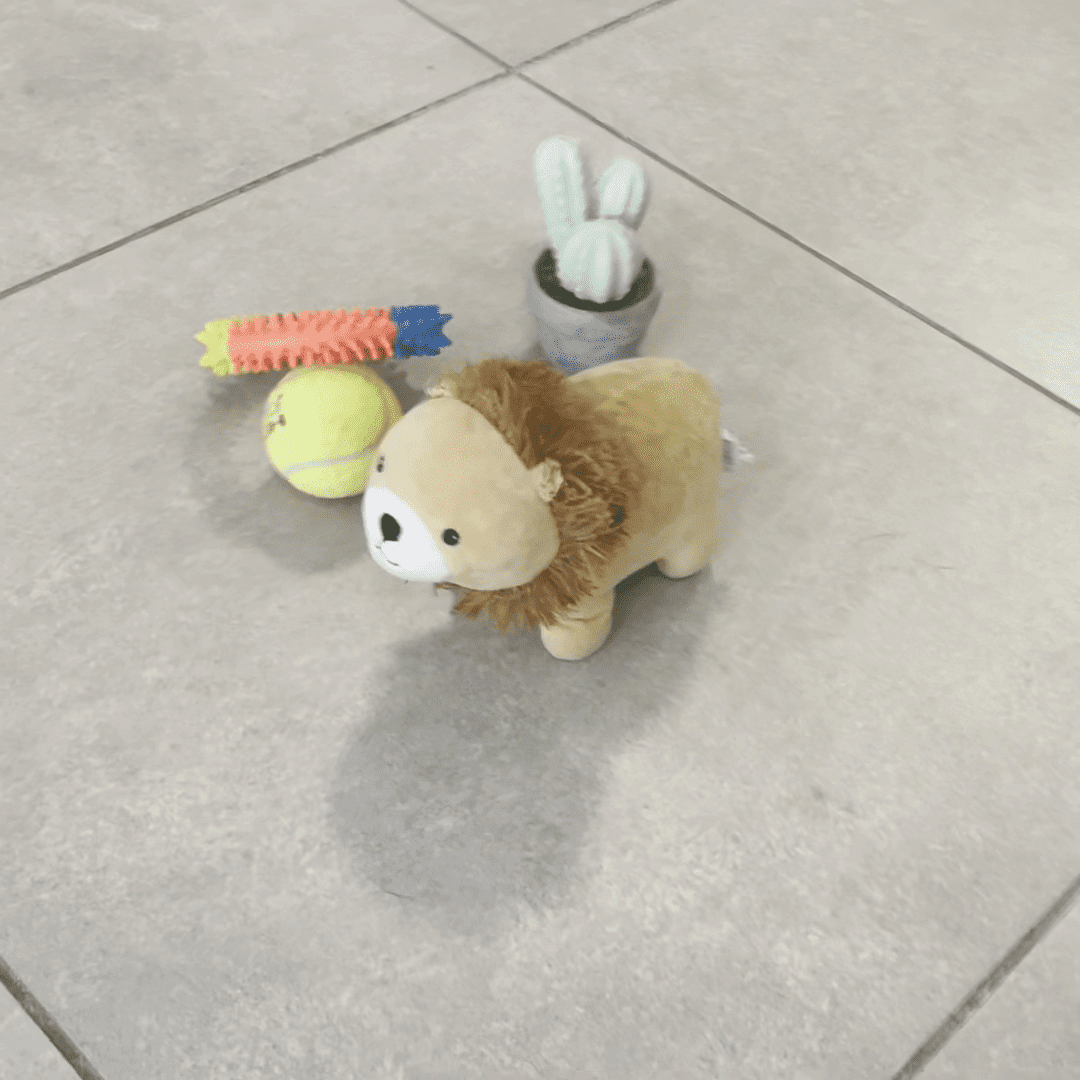}
    \end{subfigure}
    \hfill
    \begin{subfigure}[b]{0.24\linewidth}
        \includegraphics[width=\textwidth]{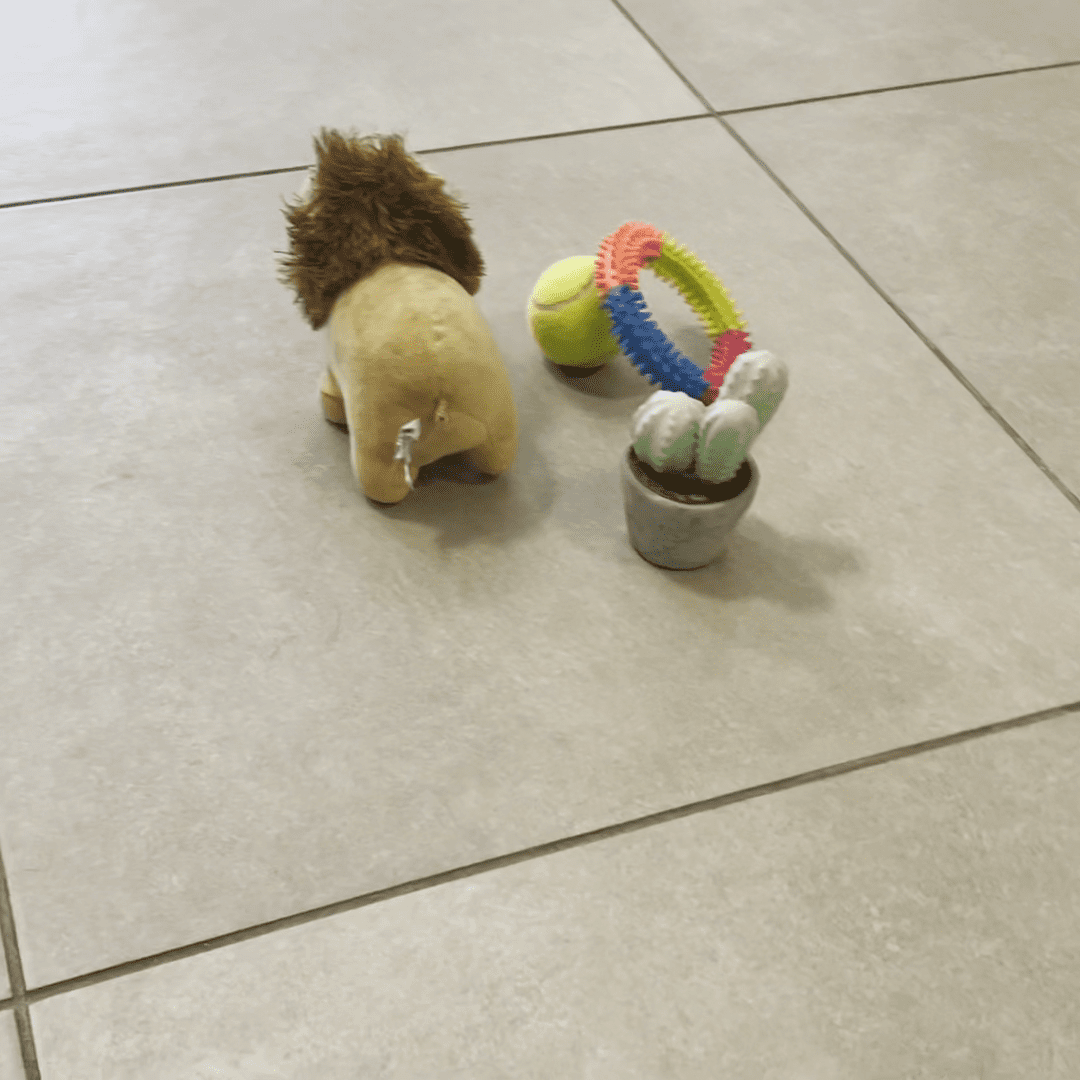}
    \end{subfigure}
    \hfill
    \begin{subfigure}[b]{0.24\linewidth}
        \includegraphics[width=\textwidth]{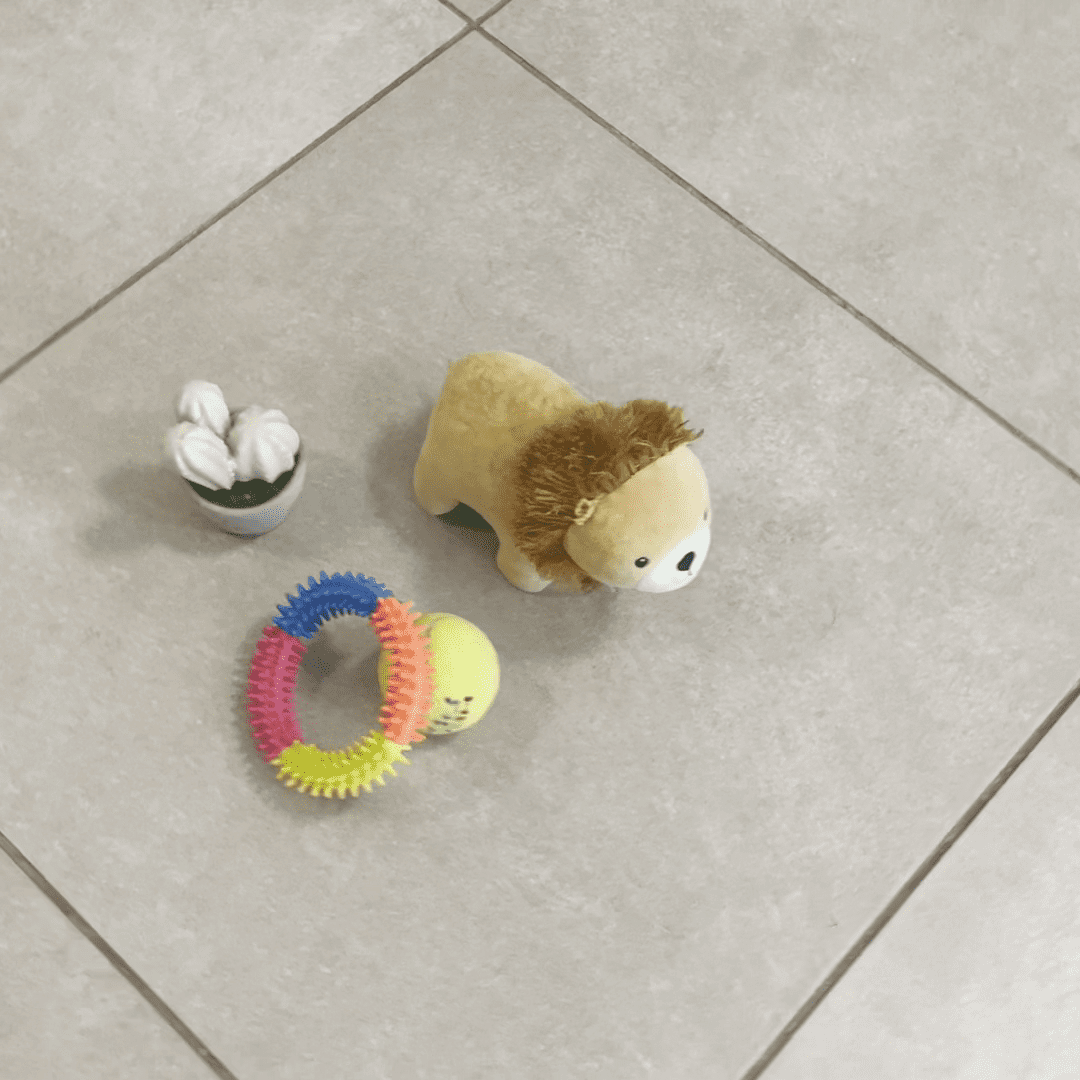}
    \end{subfigure}
    \hfill
    \begin{subfigure}[b]{0.24\linewidth}
        \includegraphics[width=\textwidth]{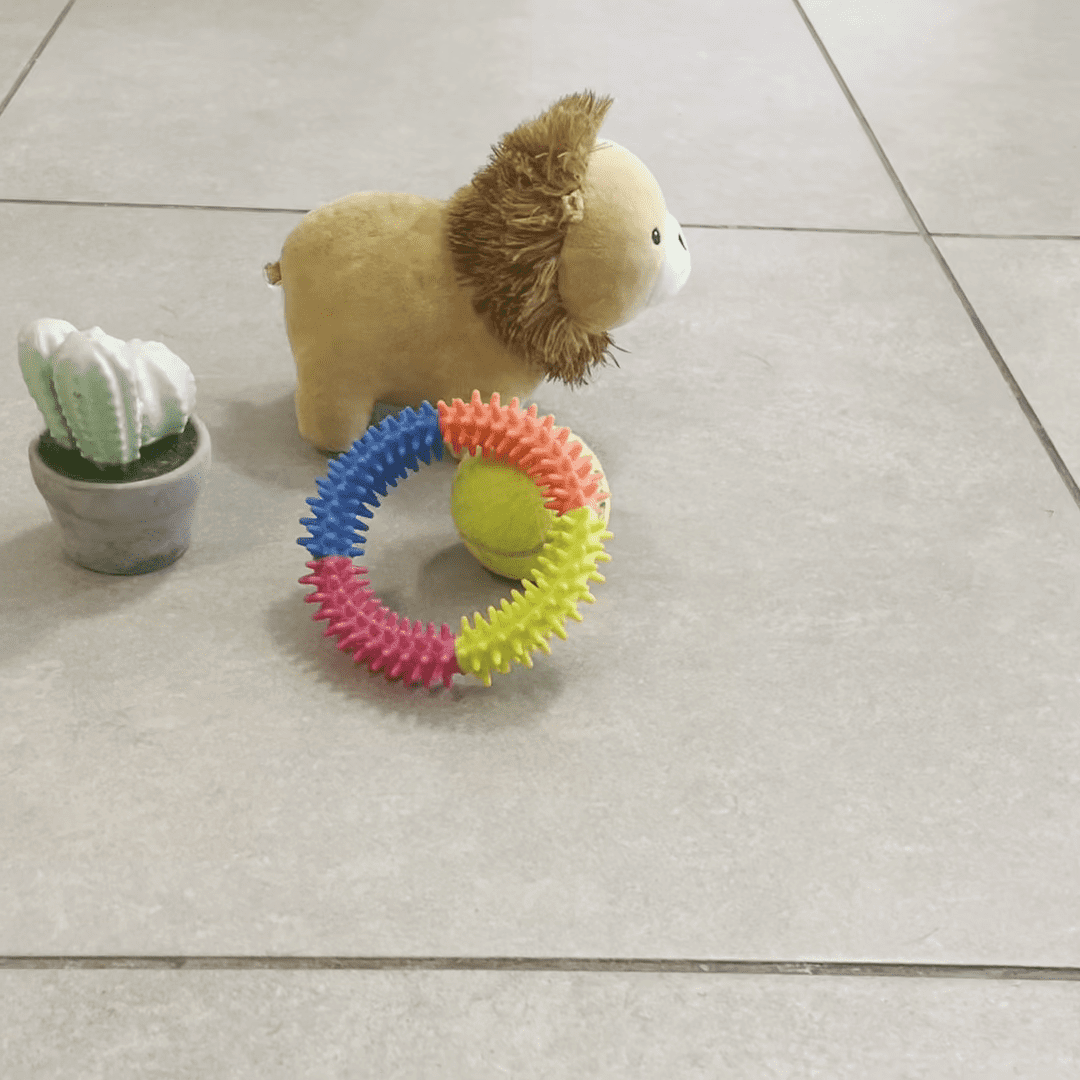}
    \end{subfigure}
    \hfill
\caption{\textbf{Lion Scene:} Four frames from the casually captured "Lion" scene}
\label{fig:lion_example}
\end{figure}

\begin{figure}[t]
\centering
    \begin{subfigure}[b]{0.24\linewidth}
        \includegraphics[width=\textwidth]{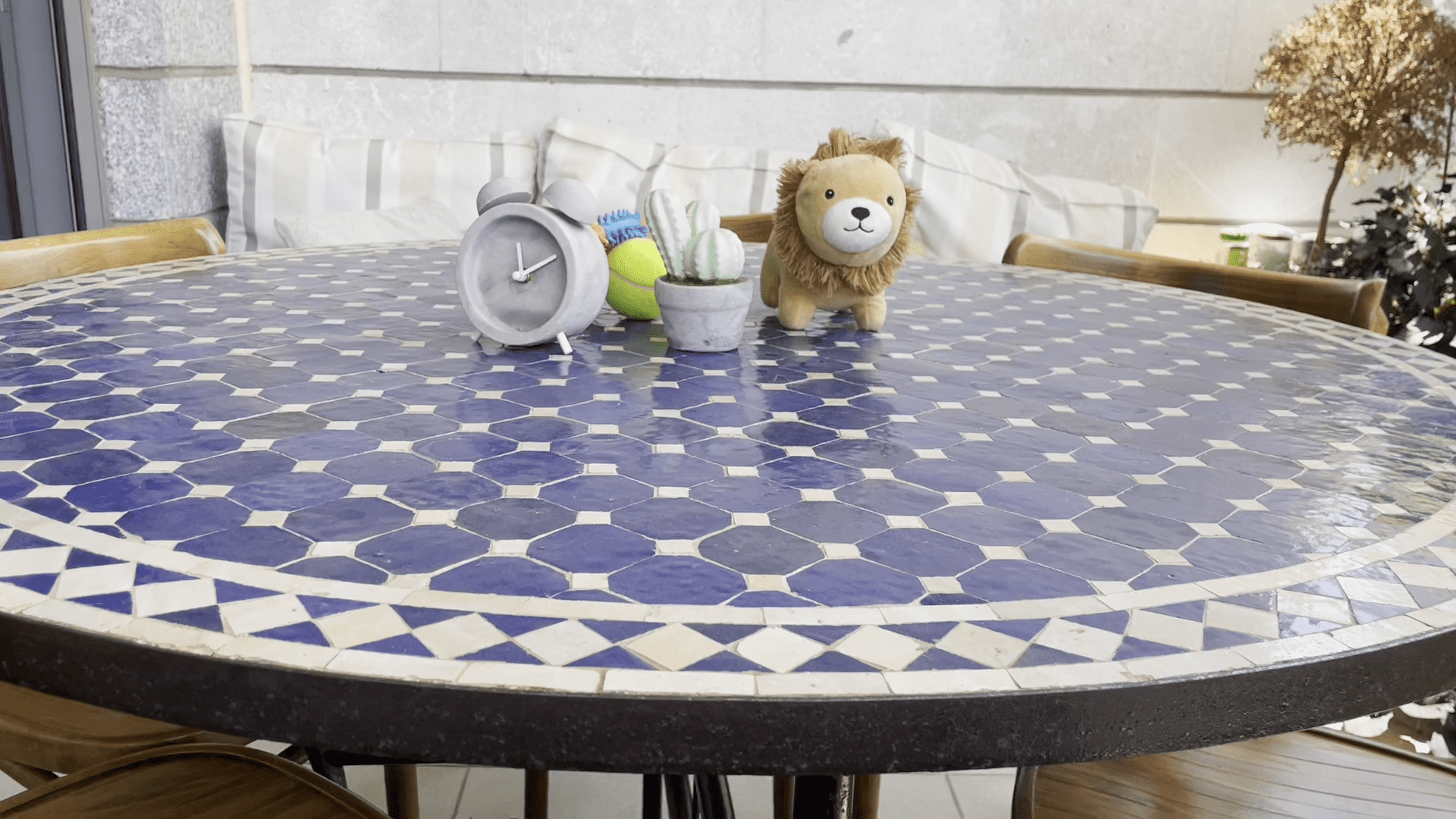}
    \end{subfigure}
    \hfill
    \begin{subfigure}[b]{0.24\linewidth}
        \includegraphics[width=\textwidth]{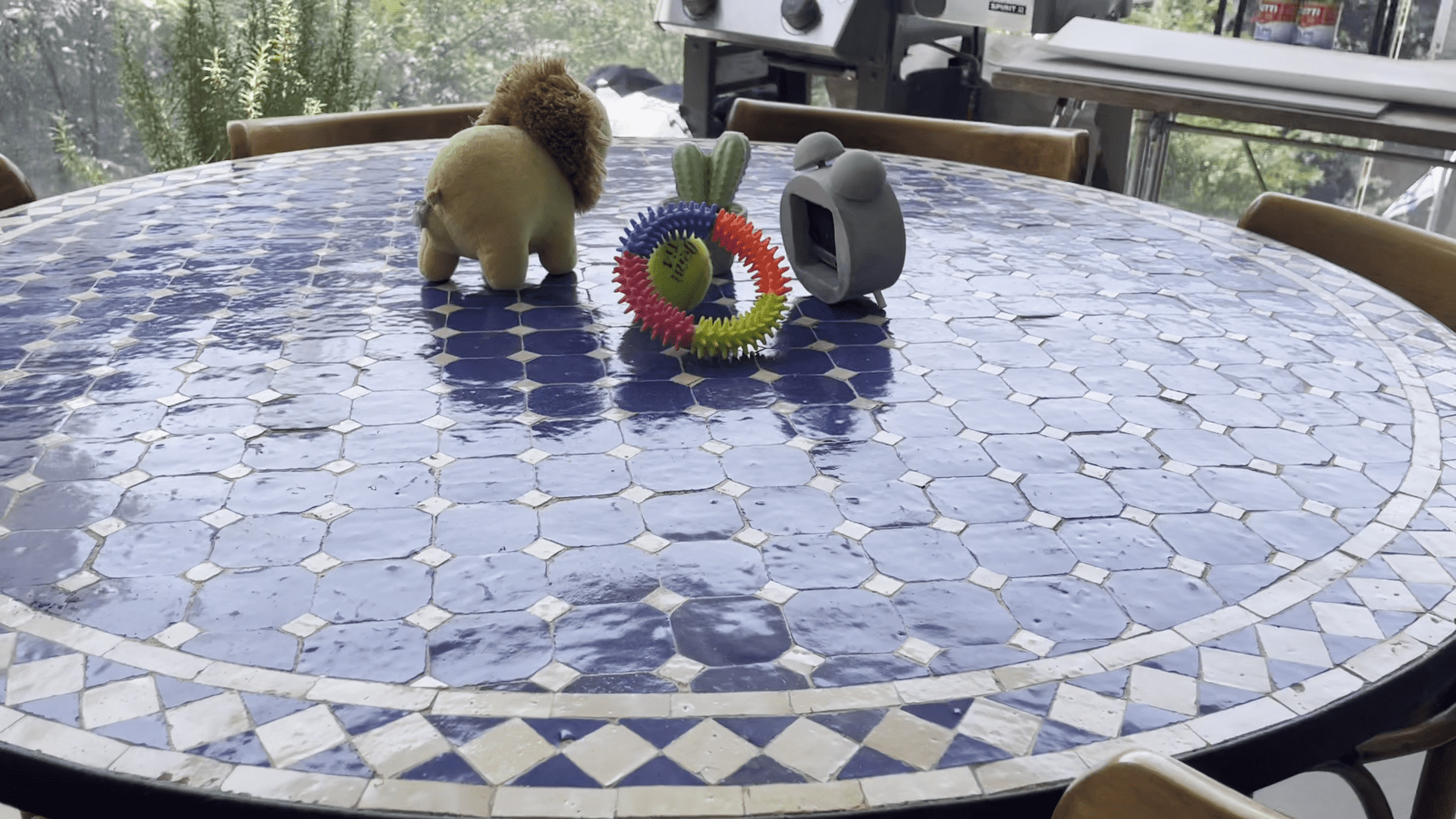}
    \end{subfigure}
    \hfill
    \begin{subfigure}[b]{0.24\linewidth}
        \includegraphics[width=\textwidth]{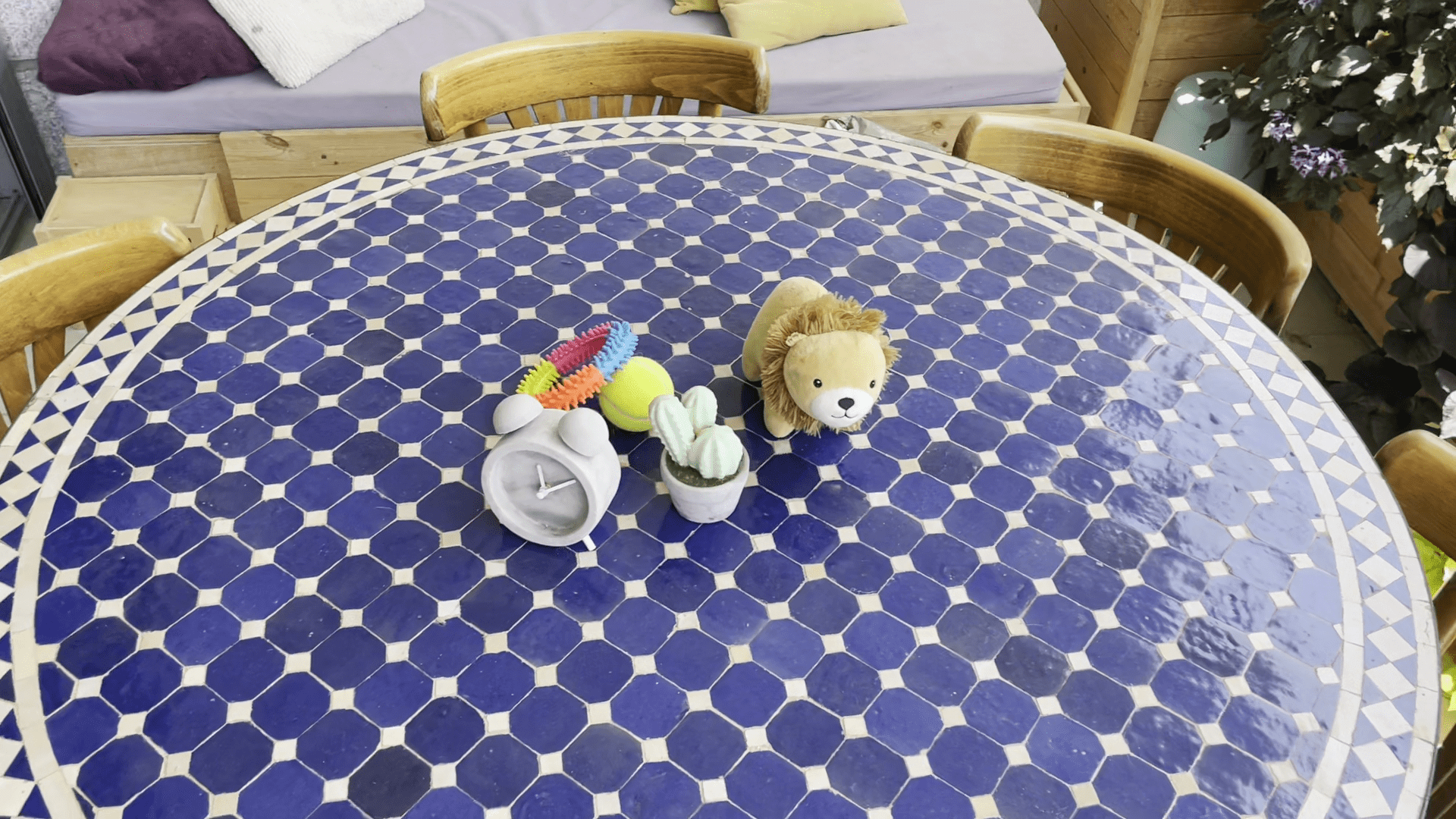}
    \end{subfigure}
    \hfill
    \begin{subfigure}[b]{0.24\linewidth}
        \includegraphics[width=\textwidth]{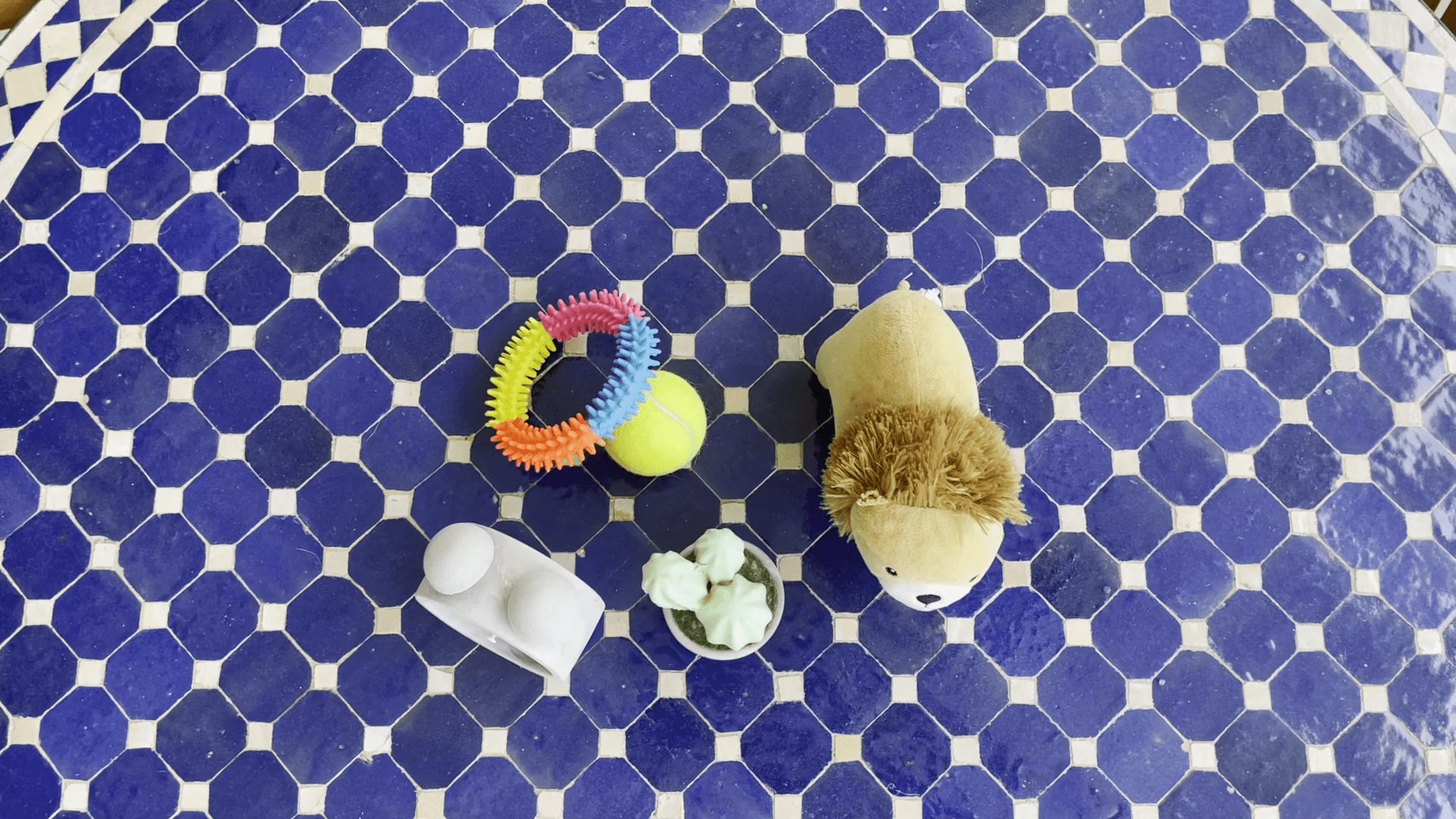}
    \end{subfigure}
    \hfill
\caption{\textbf{Table Scene:} Four frames from the casually captured "Table" scene}
\label{fig:table_example}
\end{figure}

We used our own captured scene called "casually captured" to demonstrate the registration task on naturally captured scenes. Since this data is new, we show some examples from each scene.
\Cref{fig:lion_example} shows frames from the "Lion" scene from the "Casually captured" dataset and \Cref{fig:table_example} shows frames from the "Table" scene from the "Casually captured" dataset.

\section{Implementation Details}
We used NVIDIA A5000 GPU for all the experiments, working on a single scene each time. As our NeRF representation we used Nerfacto \cite{nerfstudio}, to train NeRF we sample 1024 rays each iteration, using Adam \cite{kingma2017adam} as an optimizer with an initial learning rate of $1e^-2$ and exponential decay. 

To learn the VF we follow Real-NVP \cite{dinh2017density} with L=4 layers and H=128 hidden dims, as an optimizer we use RAdam \cite{liu2021variance} with a constant learning rate of $5e^-5$. The real scene NeRFs trained for 60K iterations and the VF train is enabled on the last 10K iterations. The synthetic scene NeRFs are trained for 20K iterations and the VF train is enabled on the last 5K iterations and ignored where the image is transparent ($\alpha < 128$ for RGBA images).

Photometric initialization is done over 25 random transformations. For PC initialization we first generate point clouds by sampling 100K points from the VF distribution, choose the points with a density higher than 10, and use these point clouds as input for the classic global registration method. In our case, we use FPFH~\cite{5152473} for feature extraction.

As for the registration phase, for the real scenes, we optimize the 6DoF parameters for 15K iterations with 32K samples per iteration, we use SGD optimizer with an initial learning rate of $5e^-3$ and exponential decay. For the synthetic scenes, we optimize the 6DoF parameters for 2.5K iterations with 8128 samples per iteration, we use SGD optimizer with an initial learning rate of $1e-3$ and exponential decay.

\section{Full Results}
\Cref{Tab:LLFF_full_results} shows the full results over LLFF dataset and \Cref{table:full_objaverse_results} shows the full results over Objaverse dataset. That is the raw transformation and rotation error for each scene and each model.
\begin{table}
\caption{
\textbf{LLFF Results:} Rotation and translation RMS error comparison on LLFF dataset, divided into the three experiment types regarding overlap between the frames. $\Delta t$ denotes the RMS translation errors multiplied by $1e2$. FPFH + RANSAC uses point clouds generated by Viewshed Fields. VF-NeRF + PC Init refers to our algorithm after initializing with VF-based points clouds using FPFH + RANSAC. VF-NeRF + Photo Init refers to our method using photometric based initialization.
}

\centering
\resizebox{0.99\textwidth}{!}{
\begin{tabular}{l | c c | c c | c c | c c }

\toprule
\multirow{2}{*}{\textbf{Model}}

&\multicolumn{2}{|c|}{\textbf{Fern}} 
&\multicolumn{2}{c|}{\textbf{Horns}}
&\multicolumn{2}{c|}{\textbf{Room}}
&\multicolumn{2}{c}{\textbf{Trex}}
\\

&\textbf{$\Delta t \downarrow$} &\textbf{$\Delta R \downarrow$} 
&\textbf{$\Delta t \downarrow$} &\textbf{$\Delta R \downarrow$}
&\textbf{$\Delta t \downarrow$} &\textbf{$\Delta R \downarrow$}
&\textbf{$\Delta t \downarrow$} &\textbf{$\Delta R \downarrow$}

\\
\midrule
\multicolumn{9}{c}{\textbf{Full Overlap}} 
\\
\midrule

FPFH~\cite{5152473} + RANSAC~\cite{10.1145/358669.358692} 
& 0.9681 & 1.3997 & 1.7415 & 1.2832 & 2.5040 & 1.9991 & 3.2461 & 1.4443
\\

iNeRF \cite{yen2020inerf}
& 8.8749 & 10.4430 & 19.2539 & 19.6978 & 38.8494 & 3.2129 & 24.5847 & 31.1075
\\

iNeRF \cite{yen2020inerf} + Photo Init.
& 0.0354 & \textbf{0.0214} & 0.0385 & \textbf{0.0055} & 0.7643 & 0.6128 & 0.0634 & \textbf{0.0098}
\\

VF-NeRF + Photo Init.
& 0.0371 & 0.0430 & 0.0107 & 0.0201 & 0.0054 & \textbf{0.0072} & \textbf{0.0073} & 0.0122
\\

VF-NeRF + PC Init. 
& \textbf{0.0321} & 0.0276 & \textbf{0.0078} & 0.0152 & \textbf{0.0045} & 0.0076 & 0.0205 & 0.0122
 \\

 \midrule
\multicolumn{9}{c}{\textbf{Partial Overlap}} 
\\
\midrule

FPFH~\cite{5152473} + RANSAC~\cite{10.1145/358669.358692} 
& 9.6166 & 5.6116 & 1.6095 & 1.6392 & 2.0541 & 1.8564 & 1.3781 & 1.7431
\\

iNeRF \cite{yen2020inerf}
& 8.3334 & 14.5468 & 11.6589 & 21.2608 & 0.0478 & 0.0199 & 61.2590 & 14.2920
\\

iNeRF \cite{yen2020inerf} + Photo Init.
& \textbf{0.0583} & \textbf{0.0235} & 51.9557 & 9.9470 & 0.0280 & 0.0248 & 0.0659 & 0.0252
\\

VF-NeRF + Photo Init.
& 0.0885 & 0.1021 & 0.0301 & 0.0219 & \textbf{0.0153} & \textbf{0.0145} & 0.0232 & \textbf{0.0045}
\\

VF-NeRF + PC Init. 
& 0.0724 & 0.0891 & \textbf{0.0179} & \textbf{0.0127} & 0.0335 & 0.0233 & \textbf{0.0189} & 0.0130
\\

 \midrule
\multicolumn{9}{c}{\textbf{No Overlap}} 
\\
\midrule

FPFH~\cite{5152473} + RANSAC~\cite{10.1145/358669.358692} 
& 3.6367 & 2.9586 & 1.0247 & 1.6241 & 3.2338 & 3.1353 & 1.1763 & 3.5331
\\

iNeRF \cite{yen2020inerf}
& 16.0154 & 13.9615 & 0.0242 & \textbf{0.0164} & 10.6372 & 12.0546 & 8.9484 & 14.3406
\\

iNeRF \cite{yen2020inerf} + Photo Init.
& 0.0678 & \textbf{0.0364} & 11.1795 & 22.5305 & 0.0435 & 0.0166 & 0.0493 & 0.0219
\\

VF-NeRF + Photo Init.
& 0.0787 & 0.1029 & 0.0325 & 0.0183 & 0.0150 & \textbf{0.0063} & \textbf{0.0035} & 0.0157
\\

VF-NeRF + PC Init. 
& \textbf{0.0544} & 0.0545 & \textbf{0.0225} & 0.0246 & \textbf{0.0123} & 0.0247 & 0.0102 & \textbf{0.0105}
\\

\bottomrule

\end{tabular}
}

\label{Tab:LLFF_full_results}
\end{table}

\begin{table*}
   \caption{Quantitative results of registration on the Objaverse dataset. $\Delta \mathbf{R}$ denotes the relative rotation errors in degree, $\Delta \mathbf{t}$ denotes the relative translation errors multiplied by 1e2 with unknown scales. $\text{DReg}_{\text{df}}$ refers to DReg with density fields and DReg refers to DReg with surface field. FPFH + RANSAC uses point clouds generated by Viewshed Fields. VF-NeRF + PC Init refers to our algorithm after initializing with VF-based points clouds using FPFH + RANSAC. The results of FGR, REGTR and DReg are taken from ~\cite{chen2023dregnerf}.} 
  \centering
  \resizebox{0.98\textwidth}{!}{
    \begin{tabular}{c | l r r r r r r r r r r r r}
      \toprule

        &  & 
        \multicolumn{1}{l}{\textbf{Food 5648}} & \multicolumn{1}{l}{\textbf{Chair 4b05}} &  
        \multicolumn{1}{l}{\textbf{Chair 4659}} & \multicolumn{1}{l}{\textbf{Chair 3f2d}} &
        \multicolumn{1}{l}{\textbf{Cone 37b5}} & \multicolumn{1}{l}{\textbf{Figurine 260d}} & 
        \multicolumn{1}{l}{\textbf{Figurine 0a5b}} & \multicolumn{1}{l}{\textbf{Figurine 09f0}} & \multicolumn{1}{l}{\textbf{Banana 3a07}} & \multicolumn{1}{l}{\textbf{Banana 2373}} & \multicolumn{1}{l}{\textbf{Banana 0a07}}
      \\
      
      \midrule

      \multirow{4}{*}{$\Delta \mathbf{R}$} 
      & FGR~\cite{zhou2016fast} & 178.34 &  50.50 &  28.54 &  81.31 & 104.52 &  89.13 &  26.35 & 138.00 &  12.17 &   6.92 &   2.86
                                           \\
      & REGTR~\cite{yew2022regtr} & 169.07 & 150.38 &  92.80 &  98.67 &  62.50 & 111.80 & 106.12 & 176.48 & 136.02 & 178.36 & 173.96
                                           \\
      & $\text{DReg}_{\text{df}}$~\cite{chen2023dregnerf} & 77.48 & 160.13 & 157.21 & 22.91 & 108.09 & 121.32 & 10.53 & 95.89 & 95.43 & 3.49 & 6.96  \\
      & DReg~\cite{chen2023dregnerf} & 6.01 &  6.53 & 17.74 & 18.88 & \textbf{18.79} & 2.11 & 7.62 & 8.25 & 15.55 & 10.95 & 1.36
                         \\
      & FPFH~\cite{5152473} + RANSAC~\cite{10.1145/358669.358692} & 3.69 & 3.11 & 1.91 & 4.06 & 54.52 & 4.31 & 4.27 & 9.80 & 2.06 & 1.69 & 1.74
                         \\
      & VF-NeRF + PC Init.    & \textbf{0.03} & \textbf{0.03} & \textbf{0.03} & \textbf{0.80} & 49.05 & \textbf{0.84} & \textbf{0.60} & \textbf{0.92} & \textbf{0.03} & \textbf{1.28} & \textbf{0.03}
      \\

      \hline
      
      \multirow{4}{*}{$\Delta \mathbf{t}$} 
      & FGR~\cite{zhou2016fast} & 17.44 &  \textbf{2.27} &  7.10 &  8.65 &  30.49 & 19.25 & 10.93 & 35.22 &  8.50 & 1.53  &  1.36
                                           \\
      & REGTR~\cite{yew2022regtr} & 30.72 & 15.41 & 24.97 & 60.53 &  84.20 & 62.07 & 35.48 & 42.10 & 10.75 & 50.40 & 13.17
                                           \\
      & $\text{DReg}_{\text{df}}$~\cite{chen2023dregnerf} & 15.52 & 7.32 & 11.72 & 2.29 & 21.70 & 33.61 & 1.95 & 21.40 & 13.14 & 4.28 & 0.50 \\
      & DReg~\cite{chen2023dregnerf} &  1.78 & 4.13 & 8.74 & 5.07 & \textbf{3.06} & 3.54 & 10.68 & 3.18 & 0.46  & 1.00 & 1.22  
                         \\
      & FPFH~\cite{5152473} + RANSAC~\cite{10.1145/358669.358692} & 0.67 & \textbf{0.42} & 3.41 & \textbf{0.45} & 23.98 & 2.82 & 0.55 & 3.16 & 0.50 & \textbf{0.13} & 3.01
                         \\
      & VF-NeRF + PC Init. & \textbf{0.16} & 2.43 & \textbf{0.05} & 1.33 & 23.73 & \textbf{0.26} & \textbf{0.005} & \textbf{0.16} & \textbf{0.27} & 0.15 & \textbf{0.44}
      \\

      \bottomrule
    \end{tabular}
  }
  
  \resizebox{0.98\textwidth}{!}{
    \begin{tabular}{c | l r r r r r r r r r r r r}
      \toprule

        &  & 
      \multicolumn{1}{l}{\textbf{Fireplug 06d5}} & \multicolumn{1}{l}{\textbf{Fireplug 0063}} & \multicolumn{1}{l}{\textbf{Fireplug 0152}} & \multicolumn{1}{l}{\textbf{Shoe 18c3}} &
      \multicolumn{1}{l}{\textbf{Shoe 1627}} & \multicolumn{1}{l}{\textbf{Shoe 0bf9}} &
      \multicolumn{1}{l}{\textbf{Shoe 022c}} & \multicolumn{1}{l}{\textbf{Teddy 1b47}} &
      \multicolumn{1}{l}{\textbf{Elephant 183a}} & \multicolumn{1}{l}{\textbf{Elephant 1608}} & \multicolumn{1}{l}{\textbf{Elephant 1a39}}
      \\
      
      \midrule

      \multirow{4}{*}{$\Delta \mathbf{R}$} 
      & FGR~\cite{zhou2016fast} & 6.19 & 20.32 &  7.50 & 10.23 & 178.14 & 71.55 & 50.28 & 8.05 & 7.65 &  21.37 & 30.97
                                           \\
      & REGTR~\cite{yew2022regtr}  & 156.92 & 99.60 & 4.04 & 2.55 & 175.21 &  97.92 &  154.91 & 149.17 & 177.15 & 172.28 & 102.62
                                            \\
      & $\text{DReg}_{\text{df}}$~\cite{chen2023dregnerf} & 156.17 & 45.76 & 12.34 & 14.69 & 131.66 & 158.66 & 6.84 & 6.32 & 6.97 & 3.92 & 126.94
                                  \\
      & DReg~\cite{chen2023dregnerf} & 7.96 & 17.43 &  4.86 & 6.06 & \textbf{12.95} & \textbf{6.48} & 2.93 & 11.44 & 8.00 & 11.13 & 13.84
                         \\
      & FPFH~\cite{5152473} + RANSAC~\cite{10.1145/358669.358692} &      5.65 & 2.06 & 1.53 & 2.07 & 14.38 & 6.44 & 2.96 & 7.14 & 1.23 & 8.30 & 1.28                   
      \\
      & VF-NeRF + PC Init. & \textbf{0.36} & \textbf{0.03} & \textbf{1.11} & \textbf{1.06} & \textbf{1.00} & \textbf{5.38} & \textbf{0.03} & \textbf{6.07} & \textbf{1.09} & \textbf{3.25} & \textbf{0.71}
      \\
      \hline
      
      \multirow{4}{*}{$\Delta \mathbf{t}$} 
      & FGR~\cite{zhou2016fast} & 5.83 &  0.83 &  1.17 &  \textbf{0.04} & 4.99 & 8.82 &  35.47 &  1.11 & 4.51 & 14.08 & 11.03
                                           \\
      & REGTR~\cite{yew2022regtr} & 68.71 & 38.74 & 2.13 & 3.53 & 43.40 & 61.37 & 102.00 & 42.84 & 52.26 & 66.15 & 34.54
                                           \\
      & $\text{DReg}_{\text{df}}$~\cite{chen2023dregnerf} & 10.54 & 5.32 & 2.60 & 4.66 & 28.63 & 24.82 & 4.40 & 2.20 & 4.26 & \textbf{1.40} & 33.57  
                                  \\
      & DReg~\cite{chen2023dregnerf} & 1.58 & 5.08 & \textbf{0.96} & 2.08 & 12.80  & 1.81 & 0.65 & \textbf{1.06} & 8.97 & 6.17 & 7.80
                         \\
      & FPFH~\cite{5152473} + RANSAC~\cite{10.1145/358669.358692} & 0.47 & 0.17 & 2.98 & 1.33 & 7.40 & \textbf{1.08} & 2.05 & 6.54 & 0.30 & 3.88 & 0.86
                         \\
      & VF-NeRF + PC Init. & \textbf{0.006} & \textbf{0.023} & 2.60 & 0.24 & \textbf{0.13} & 4.56 & \textbf{0.19} & 5.90 & \textbf{0.17} & 2.57 & \textbf{0.76}
\\
      \bottomrule
    \end{tabular}
  }



   
   \resizebox{0.98\textwidth}{!}{
     \begin{tabular}{c | l r r r r r r r r r r r r}
       \toprule
 
         &  & 
         \multicolumn{1}{l}{\textbf{Piano 0e0d}} & \multicolumn{1}{l}{\textbf{Piano 0a6e}} &
         \multicolumn{1}{l}{\textbf{Truck 1431}} & \multicolumn{1}{l}{\textbf{Guitar 15b4}} & \multicolumn{1}{l}{\textbf{Guitar 14f8}} & \multicolumn{1}{l}{\textbf{Guitar 0ceb}} &
         \multicolumn{1}{l}{\textbf{Guitar 0aa0}} & \multicolumn{1}{l}{\textbf{Lantern 0231}} & \multicolumn{1}{l}{\textbf{Lamp 0230}} & \multicolumn{1}{l}{\textbf{Bench 0b05}} & \multicolumn{1}{l}{\textbf{Shield 22a7}} 
       \\
       
       \midrule
 
       \multirow{4}{*}{$\Delta \mathbf{R}$} 
       & FGR~\cite{zhou2016fast} & 23.09 & 77.63 & 7.46 & 7.80 & 5.25 & 13.07 & 39.94 & 130.36 & 17.44 & 19.51 & 170.27   
                                            \\
       & REGTR~\cite{yew2022regtr} & 30.54 & 117.90 & 178.49 & 5.18 & 29.47 & 103.84 & 5.95 & 139.32 & 160.45 & 122.12 & 157.38
                                            \\
       & $\text{DReg}_{\text{df}}$~\cite{chen2023dregnerf} & 160.71 & 168.79 & 117.07 & 11.43 & 164.87 & 177.62 & 7.96 & 7.76 & 173.09 & 179.16 & 178.00 \\
       & DReg~\cite{chen2023dregnerf} & 16.30 & 13.51 & 16.68 & 12.60 & 3.43 & \textbf{1.08} & 9.53 & 9.17 & 16.44 & 12.98 & 8.21
                          \\
      & FPFH~\cite{5152473} + RANSAC~\cite{10.1145/358669.358692} & \textbf{5.80} & 12.86 & 6.62 & 1.84 & 3.91 & 2.06 & 2.44 & 2.69 & 2.82 & 2.47 & \textbf{1.29}              
      \\
      & VF-NeRF + PC Init. & 6.89 & \textbf{6.37} & \textbf{0.03} & \textbf{1.34} & \textbf{1.98} & 1.16 & \textbf{1.54} & \textbf{0.83} & \textbf{0.03} & \textbf{0.03} & 3.68
      \\
       \hline
       
       \multirow{4}{*}{$\Delta \mathbf{t}$} 
       & FGR~\cite{zhou2016fast} & 7.43 & 14.50 & 5.95 & 2.86 & 3.46 & 1.83 & 8.42 & 9.06 & 0.69 & 12.52 & 15.57
                                            \\
       & REGTR~\cite{yew2022regtr} & 44.24 & 65.99 & 50.63 & 15.18 & 18.41 & 89.20 & 9.91 & 57.25 & 64.44 & 31.97 & 44.29
                                            \\
       & $\text{DReg}_{\text{df}}$~\cite{chen2023dregnerf} & 22.86 & 26.18 & 24.83 & 10.27 & 8.50 & 43.21 & 4.09 & 3.35 & 26.77 & 28.59 & 34.71 \\
       & DReg~\cite{chen2023dregnerf} & 4.80 & 12.54 & \textbf{0.04} & 5.72 & 5.03 & 3.01 & 1.20 & 3.29 & 1.31 & 1.68 & 11.33
                          \\
       & FPFH~\cite{5152473} + RANSAC~\cite{10.1145/358669.358692} & \textbf{3.74} & \textbf{3.14} & 4.27 & 2.15 & 1.05 & 1.47 & 1.16 & 2.43 & 0.91 & 5.43 & \textbf{3.07}
                         \\
      & VF-NeRF + PC Init. & 13.84 & 6.45 & 0.36 & \textbf{0.14} & \textbf{0.14} & \textbf{0.74} & \textbf{0.16} & \textbf{0.36} & \textbf{0.11} & \textbf{0.27} & 8.03
      \\
       \bottomrule
     \end{tabular}
   }
   
   \resizebox{0.98\textwidth}{!}{
     \begin{tabular}{c | l r r r r r r r r r r r r}
       \toprule
 
         &  & 
       \multicolumn{1}{l}{\textbf{Shield 1973}} & \multicolumn{1}{l}{\textbf{Shield 14a6}} &
       \multicolumn{1}{l}{\textbf{Shield 00ad}} & \multicolumn{1}{l}{\textbf{Controller 0866}} & \multicolumn{1}{l}{\textbf{Fighter Jet 16c6}} & \multicolumn{1}{l}{\textbf{Fighter Jet 089f}} & \multicolumn{1}{l}{\textbf{Fighter Jet 0000}} & \multicolumn{1}{l}{\textbf{Telephone 1a8c}} & \multicolumn{1}{l}{\textbf{Telephone 0354}} & \multicolumn{1}{l}{\textbf{Lampshade ab66}} & \multicolumn{1}{l}{\textbf{Skateboard 10c7}} 
       \\
       
       \midrule
 
       \multirow{4}{*}{$\Delta \mathbf{R}$} 
       & FGR~\cite{zhou2016fast} & 130.83 & 178.99 & 7.06 & 164.01 & 11.91 & 40.21 & 39.86 & 150.10 & 19.76 & 147.50 & 176.90
                                            \\
       & REGTR~\cite{yew2022regtr}  & 138.94 & 169.78 & 14.76 & 102.05 & 154.74 & 150.64 & 178.35 & 144.47 & 1.13 & 148.44 & 3.88
                                             \\
       & $\text{DReg}_{\text{df}}$~\cite{chen2023dregnerf} & 178.93 & \textbf{4.92} & 7.78 & 179.13 & 9.75 & 23.97 & 178.68 & 132.49 & 16.59 & 5.79 & 179.97 \\
       & DReg~\cite{chen2023dregnerf} & 12.94 & 12.26 & 2.29 & 4.03 & 6.88 & 10.53 & 6.46 & 15.60 & 9.01 & 5.67 & \textbf{1.92}
                          \\
       & FPFH~\cite{5152473} + RANSAC~\cite{10.1145/358669.358692} & 3.92 & 13.62 & \textbf{1.14} & 7.14 & 1.50 & 3.55 & \textbf{1.31} & 23.12 & 3.32 & 1.98 & 179.45
                         \\
      & VF-NeRF + PC Init. & \textbf{2.74} & 13.46 & 1.47 & \textbf{0.03} & \textbf{0.26} & \textbf{0.03} & 2.12 & \textbf{0.03} & \textbf{0.03} & \textbf{1.19} & 179.10
      \\
       \hline
       
       \multirow{4}{*}{$\Delta \mathbf{t}$} 
       & FGR~\cite{zhou2016fast} & 49.44 & 55.62 & \textbf{0.22} & 8.97 & 6.90 & 11.73 & 3.44 & 57.53 & 18.51 & 67.08 & 19.03
                                            \\
       & REGTR~\cite{yew2022regtr} & 72.01 & 48.56 & 6.57 & 65.87 & 52.97 & 82.10 & 28.79 & 51.09 & 0.91 & 54.45 & 5.14
                                            \\
       & $\text{DReg}_{\text{df}}$~\cite{chen2023dregnerf} & 59.08 & \textbf{0.66} & 1.58 & 23.68 & 2.64 & 13.98 & 19.28 & 63.76 & 15.65 & 4.81 & 12.20 \\
       & DReg~\cite{chen2023dregnerf} & 2.79 & 4.38 & 1.26 & \textbf{0.99} & 2.53 & 7.55 & 2.09 & 1.28 & 3.57 & \textbf{0.81} & \textbf{0.16}
                          \\
       & FPFH~\cite{5152473} + RANSAC~\cite{10.1145/358669.358692} & \textbf{1.52} & 4.40 & 2.65 & 4.12 & 1.18 & 3.97 & 4.10 & 8.48 & 1.03 & 3.07 & 3.04
                         \\
      & VF-NeRF + PC Init. &  1.61 & 6.30 & 2.65 & 1.96 & \textbf{0.15} & \textbf{0.19} & \textbf{0.34} & \textbf{0.26} & \textbf{0.65} & 1.45 & 1.93
      \\
       \bottomrule
     \end{tabular}
   }

   \label{table:full_objaverse_results}
\end{table*}

\par\vfill\par

{\small
\bibliographystyle{splncs04}
\bibliography{egbib}
}

\end{document}